\documentclass[letterpaper]{article} 
\usepackage{aaai2026}  
\usepackage{times}  
\usepackage{helvet}  
\usepackage{courier}  
\usepackage[hyphens]{url}  
\usepackage{graphicx} 
\usepackage{natbib}  
\usepackage{caption} 
\usepackage{algorithm}
\usepackage{algorithmic}

\usepackage{multirow}
\usepackage{etoolbox}
\usepackage{array}
\usepackage{xcolor}
\usepackage{soul}

\newcommand{\insertfig}{\includegraphics[width=0.85\linewidth]{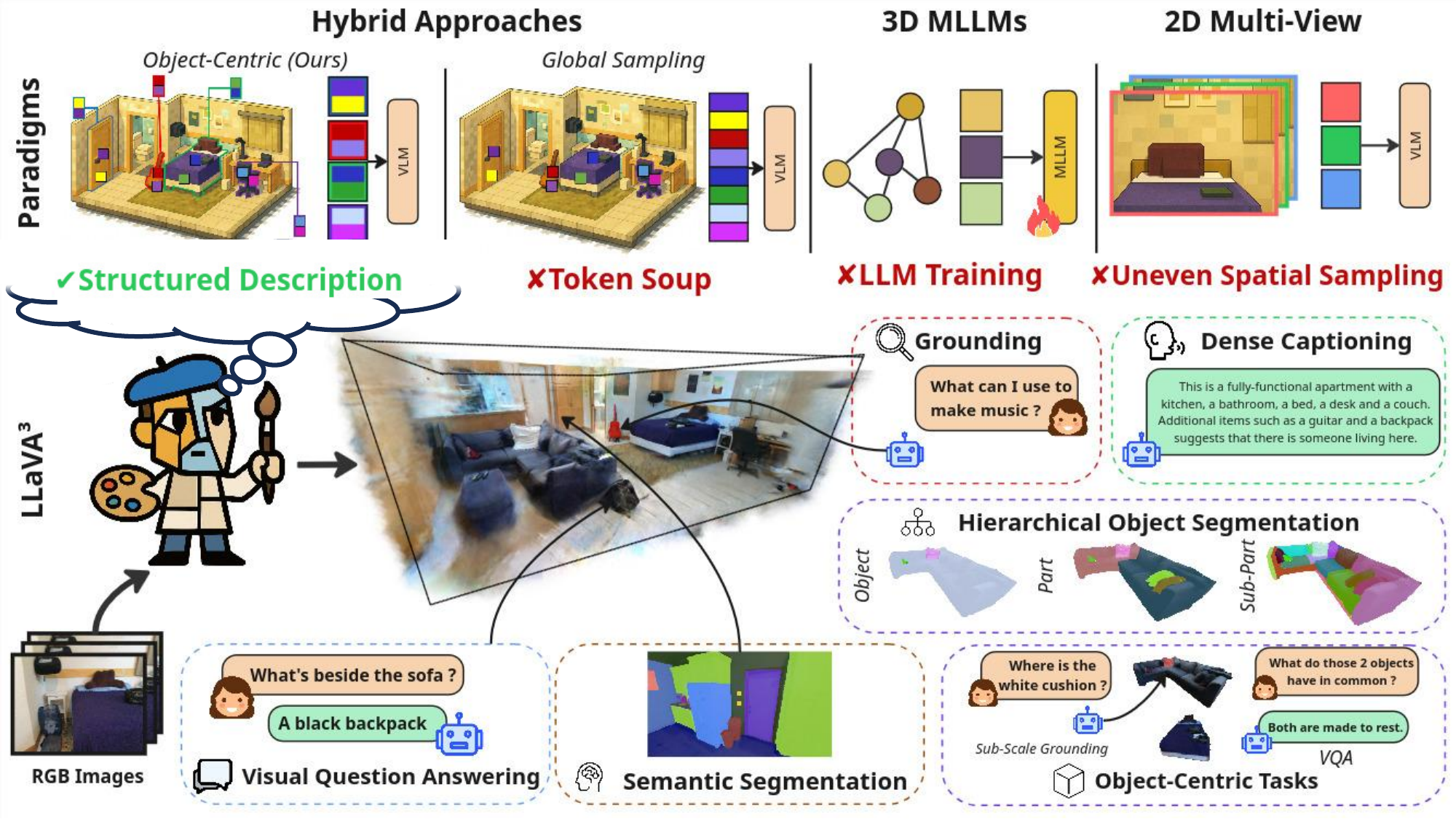}\captionof{figure}{
        \textbf{LLaVA$^3$}  empowers 3D understanding ability of Vision Language Models (VLM) through a new paradigm of 2D visual representations of 3D scenes. Our representation relies on an object-centric description of the scene, each object being visually described from a multitude of viewpoints jointly. Reconstructed from multi-view images, this representation permits VLMs to achieve various tasks such as 3D Visual Question Answering, 3D Grounding or 3D Semantic Segmentation.
}\label{fig:overview}}
\makeatletter
\apptocmd{\@maketitle}{\centering\insertfig}{}{}
\makeatother
\usepackage{newfloat}
\usepackage{listings}
\usepackage{pifont}%
\usepackage{arydshln}
\DeclareCaptionStyle{ruled}{labelfont=normalfont,labelsep=colon,strut=off} 
\floatstyle{ruled}
\newfloat{listing}{tb}{lst}{}
\floatname{listing}{Listing}
\twocolumn

\title{LLaVA$^3$ : Representing 3D Scenes like a Cubist Painter to Boost 3D Scene Understanding of VLMs} 
\author{
    Doriand Petit\textsuperscript{\rm 1,2}, Steve Bourgeois\textsuperscript{\rm 1}, Vincent Gay-Bellile\textsuperscript{\rm 1}, Florian Chabot\textsuperscript{\rm 1}, Loïc Barthe\textsuperscript{\rm 2}\\
}
\affiliations{
        \textsuperscript{\rm 1} Université Paris-Saclay, CEA, List, F-91120, Palaiseau, France\\
    \textsuperscript{\rm 2}IRIT, Université de Toulouse, CNRS, France\\
    first.last@cea.fr, first.last@irit.fr
}

\begin{document}

\maketitle

\begin{abstract}
    Developing a multi-modal language model capable of understanding 3D scenes remains challenging due to the limited availability of 3D training data, in contrast to the abundance of 2D datasets used for vision-language models (VLM). As an alternative, we introduce LLaVA³ (pronounced LLaVA Cube), a novel method that improves the 3D scene understanding capabilities of VLM using only multi-view 2D images and without any fine-tuning. Inspired by Cubist painters, who represented multiple viewpoints of a 3D object within a single picture, we propose to describe the 3D scene for the VLM through omnidirectional visual representations of each object.
    These representations are derived from an intermediate multi-view 3D reconstruction of the scene. Extensive experiments on 3D VQA and 3D language grounding show that our approach outperforms previous 2D-based VLM solutions.
\end{abstract}

\begin{links}
    \link{Webpage}{https://cea-list.github.io/LLaVA-Cube/}
\end{links}

\section{Introduction}
\label{sec:intro}

3D scene understanding is a central goal in computer vision, with broad applications in areas such as robotics~\cite{ni2023deep,naseer2018indoor} and autonomous navigation~\cite{guo2021survey}. It involves a wide range of downstream tasks, from spatial reasoning and object decomposition to dense captioning and segmentation, covering both textual and pixel-level outputs, as illustrated in Figure \ref{fig:overview}.

Recent Vision-Language Models (VLMs) such as LLaVA~\cite{liu2023llava,zhang2024vision} have demonstrated impressive capabilities in interpreting 2D images through multi-modal understanding and autoregressive language generation, enabling tasks like Visual Question Answering (VQA). Motivated by their success, multiple works have explored extending these models to 3D scene understanding. 
Specifically, 3D Multi-modal Large Language Models (3D MLLMs) are trained to directly process raw point clouds, whereas VLMs were extended to interpret multiple 2D images of the same scene at once.
Yet, despite encouraging results, 3D performance still falls short of the rich reasoning capabilities achieved in 2D. Regarding VLMs, this gap is partly due to the inherent complexity of 3D reasoning, which requires integrating information across multiple viewpoints, spatial scales, and occlusions. Regarding 3D MLLMs, the main difficulty is related to the scarcity of 3D data to train such models.

In this paper, we propose to improve the 3D scene understanding ability of VLM from multi-view images by computing a novel visual 2D scene description that better captures the 3D nature of the scene. Inspired by works from ChatSplat~\cite{chen2024chatsplat} and SplatTalk~\cite{thai2025splattalk}, our approach first achieves a multi-view reconstruction of the scene in 3D, including a 3D field of LLaVA visual tokens. This 3D field is then used to provide 2D visual descriptions of the scene. However, we observe that those initial approaches sample tokens across the scene in a spatially unstructured manner, leading to redundant or inconsistent representations and limiting VLM performance.

To address this, we introduce a structured, object-centric approach inspired by the principles of Cubist art, which deconstructs 3D objects into 2D projections.
We first segment the scene hierarchically into objects and compute an omni-directional visual description for each object by sampling from the LLaVA field. These per-object token sets are semantically rich, spatially diverse, and context-aware, and can be fed as images to a frozen LLaVA model. Objects are ordered according to their spatial layout, providing a consistent and interpretable input for downstream reasoning.

Our evaluations demonstrate that LLaVA$^3$ outperforms other VLM-based solutions on 3D Visual Question Understanding (3D VQA) while allowing many 3D downstream tasks (Figure~\ref{fig:overview}) without requiring any VLM fine-tuning.

\noindent To summarize, our contributions are:
\begin{enumerate}

\item We introduce the new concept of representing a 3D scene for a VLM as a collection of omni-directional visual-description of each of the 3D objects it contains.
\item We design the first objects hierarchy  decomposition of a 3D Neural Field scene, facilitating the interpretation of the scene due to its discrete nature.
\item We propose to model jointly view-independent and view-dependent information contained in the high-dimension LLaVA visual tokens to better capture both objects semantics and their spatial relationships.

\item  We demonstrate that LLaVA$^3$ outperforms other VLM-based solutions in term of 3D VQA and 3D grounding.
\end{enumerate}

\section{Related Works}
\label{sec:sota}

\textbf{Large Multi-modal Models for 3D Scene Understanding.} Large Language Models (LLMs) such as GPT-4~\cite{achiam2023gpt}, Claude~\cite{claude3}, and LLaMA~\cite{touvron2023llama} have established powerful foundations for natural language understanding, leading to the emergence of Large Multi-modal Models (LMMs) that extend language capabilities to visual domains. LLaVA~\cite{liu2023llava} (and its subsequent variants~\cite{liu2023improvedllava,liu2024llavanext,li2024llava}) represents a seminal work in this direction, demonstrating that LLMs can understand and reason about visual content through a simple vision encoder-language model architecture connected by a projection layer. 

Approaches to extends VLM abilities to 3D scene understanding can be grouped into several categories. First, 2D VLMs already exhibit impressive multi-view reasoning capabilities, often achieving surprisingly accurate 3D scene understanding from 2D inputs. However, these methods face key limitations: the restricted context window of language models and the mismatch between 2D image inputs and the inherently 3D nature of scenes. While recent works have sought to extend these architectures to 3D point clouds~\cite{hong20233d,guo2023point,xu2024pointllm}, adapting VLMs to point clouds remains difficult as point clouds are less informative than multi-view images and are harder to collect at scale, particularly for large-scale model training.
A growing class of hybrid methods~\cite{chen2024chatsplat,thai2025splattalk} addresses these issues by reconstructing the 3D scene and LLaVA feature field from multi-view images using NeRF or Gaussian Splatting, then summarizing the scene for the VLM via global sampling of the LLaVA feature field. This approach combines strengths of both paradigms: it leverages 3D reconstruction for dense spatial coverage while avoiding the need to train or fine-tune the VLM. Our work falls within this category, but we argue that existing sampling strategies, which usually consist in unstructured sampling across the scene,  are sub-optimal. Instead, we decompose the scene into objects and construct per-object feature sets tailored for VLM consumption. \\
\textbf{NeRF and Feature Fields.}
Neural Fields~\cite{mildenhall2021nerf} represent scenes using 3D feature grids~\cite{muller2022instant} and neural networks. They are learnt from multi-view images and associated camera poses. 
Recent works have distilled pretrained image encoders into 3D feature fields to enable open-vocabulary understanding and semantic segmentation. Methods like LeRF~\cite{kerr2023lerf}, OpenNeRF~\cite{engelmann2024opennerf} or Decomposing-NeRF~\cite{kobayashi2022decomposing} ground open-vocabulary embeddings (respectively CLIP~\cite{radford2021learning}, OpenSeg~\cite{ghiasi2022scaling} and LSeg~\cite{li2022languagedriven}) inside NeRF models, allowing textual querying across the scene. Other approaches leverage the Segment Anything Model (SAM)~\cite{kirillov2023segment}: SAM-NeRF projects SAM features into the scene for easy novel view segmentation, while Garfield~\cite{kim2024garfield} rather proposes a continuous multi-scale scene decomposition by distilling masks using contrastive learning. 
Inspired by these ideas, LLaVA$^3$ learns two complementary 3D feature fields. The first one is aligned with LLaVA for semantic reasoning. Unlike previous works, it reconstructs the view-dependency of the LLaVA token to better capture the spatial relationships among the scene elements. The second field is aligned with SAM masks and CLIP features for scene decomposition.
Unlike Garfield, our decomposition provides a discrete objects hierarchy that is easier to exploit than continuous volume-based decomposition.

\begin{figure*}[t]
    \includegraphics[width=\textwidth]{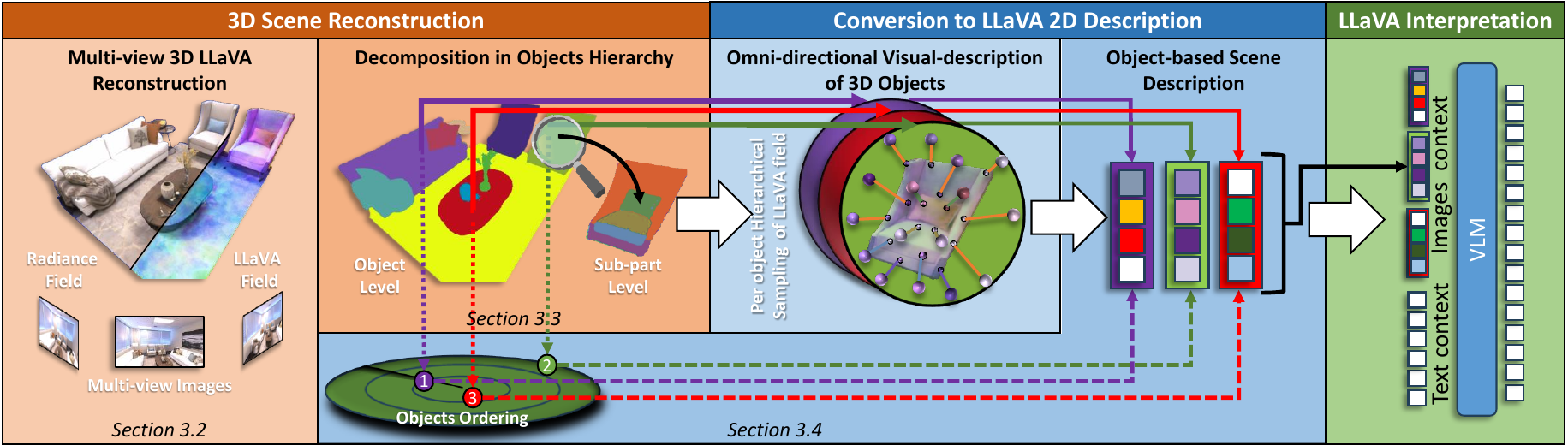}
    \caption{\textbf{Overview of LLaVA$^3$.} We first reconstruct the 3D scene as a NeRF from multi-view images with an associated LLaVA feature field. We also derive a hierarchical 3D segmentation of our NeRF. For each object, we create an omni-directional visual-description as a set of tokens. After object re-ordering, we can finally feed them to the VLM for 3D interpretation.}
    \label{fig:overview_method}
\end{figure*}

\section{Method}
\label{sec:method}
\label{sec:overview}

In this paper, we propose to improve the VLM ability to interpret 3D scenes from multiple views without any VLM fine-tuning. Instead of directly providing the visual-description of these multiple 2D views to the VLM, our solution provides an omni-directional visual-description for each object of the scene. 
As illustrated in Figure \ref{fig:overview_method}, our process first achieves a 3D grid-based NeRF reconstruction of the scene, including a LLaVA feature field (section \ref{sec:met1}). The reconstruction is then hierarchically decomposed into object/part/sub-parts, resulting into an explicit 3D object decomposition graph (section \ref{sec:met2}). An omnidirectional 2D visual-description of each object is computed by sampling tokens from the LLaVA-field equally amongst the object sub-components. This visual description is also designed to capture the object semantics while preserving the maximum of contextual information related to its relationships with the rest of the environment. Those objects' omnidirectional visual-descriptions are then provided as image tokens to the VLM, following an order depending on the object position in the 3D scene  (section \ref{sec:description}).

\subsection{Preliminaries}
\label{sec:preli}
\textbf{Neural Fields} \cite{mildenhall2021nerf} (NeRFs) are learnable neural networks (possibly coupled with multi-resolution feature hashgrids \cite{muller2022instant}) over-fitted to individual scenes, which output density $\sigma$ and color $c$ from any 3D position and view direction queries. A 2D pixel color $\hat{C}$ is recovered by sampling points along a ray cast from the corresponding posed image and compositing them via volume rendering: $\hat{C}(r) = \sum_{i=0}^{N-1} w_i c_i$, with $w_i = T_i(1-exp(-\sigma_i \delta_i))$ (which we denote as the density weights) and $T_i = exp(-\sum_{j=0}^{i-1}\sigma_j \delta_j)$, $c_i$ is the color of sample $i$ and $\delta$ is the distance between consecutive samples. The scene is optimized by minimizing the MSE loss $\mathcal{L}_{rgb} = ||\hat{C}(r) - C(r)||^2$ between rendered and ground truth colors.
Feature fields are trained similarly by replacing RGB color with d-dimensional features, optimizing the model via comparison between NeRF rendered features and feature maps from pre-trained image encoder. Multiple feature fields can be learnt on the same model, using one decoder per feature plugged into joint or separate sets of grids. \\
\textbf{LLaVA-OV}~\cite{li2024llava} integrates a SigLip~\cite{zhai2023sigmoid} image encoder $g_{\psi}$ with an LLM $f_{\phi}$, connected via an MLP projector $p_{\theta}$ that maps visual features into the language embedding space. The projected visual tokens are concatenated with textual tokens and jointly processed by the language model to produce textual outputs.

\subsection{Reconstructing a LLaVA Feature Field}
\label{sec:met1}

Our first objective is to reconstruct a dense  3D representation of the scene, including both geometry and LLaVA visual embeddings. To this end, we extend usual NeRF approach \cite{muller2022instant} with an additional LLaVA field. 
Following standard practice, we extract LLaVA token maps from training images and distill them into a 3D feature field. Despite LLaVA’s low-resolution feature maps ($27 \times 27$), NeRF’s multi-view consistency enables recovery of higher-resolution features, as shown in LeRF \cite{kerr2023lerf}. Based on ChatSplat~\cite{chen2024chatsplat} and SplatTalk~\cite{thai2025splattalk} insights, we supervise the field with post-projector features ($ p_{\theta} \circ g_{\psi}$) rather than SigLIP features. This choice stems from the projector’s high sensitivity to reconstruction noise in SigLIP features, resulting in unintelligible tokens, while supervising with post-projector outputs offers more stability by operating directly in the VLM’s token space. \\
\textbf{Reconstruction of high-dimensional feature field.}
LLaVA tokens pose a unique challenge: their high dimensionality (3584D), sparsity, and lack of normalization make them difficult to learn reliably within a compact feature field without losing their semantic integrity. SplatTalk, based on Gaussian Splatting, addresses this by training a scene-specific auto-encoder that compresses tokens into a lower-dimensional, normalized space to stabilize training. This step is necessary in GS, where high-dimensional tokens cannot be directly stored or rendered due to memory limitations~\cite{qin2024langsplat,shi2024language}. However, it introduces information loss through two stages of imperfect compression (first via the auto-encoder, then during field learning) and adds additional pre-training overhead. In contrast, because we operate within a NeRF representation, we are not bound by these limitations and can learn full-resolution LLaVA features end-to-end using an auto-decoder architecture, preserving their full expressivity. Specifically, the feature field is decoded into  normalized lower-dim feature $f$ before being mapped to the full token dimensionality $t$. \\
\textbf{Reconstructing semantics and spatial relationships.}
Visuo-language features, such as CLIP, SigLIP or LLaVA features, encode both the semantics of individual objects and their spatial relationships. Because each viewpoint offers only a partial observation of the scene, the spatial information captured in these features is inherently view-dependent, while the object-level semantics remain view-independent. \\
To our knowledge, existing multi-view 3D reconstruction methods for feature fields typically rely on view-independent modeling (i.e. assigning a single static embedding to each 3D point, regardless of the viewpoint). This approach effectively captures the object’s semantics while averaging out viewpoint-specific spatial relationships. Such filtering is well-suited for tasks like semantic segmentation, where object identity alone suffices. However, this becomes questionable in the context of vision-language models (VLMs) used for 3D scene analysis. In such cases, averaging view-dependent features can erase critical view-specific cues like inter-object relationships that are visible from very few input images.
On the other hand, view-independent features benefit from multi-view aggregation, making them better at preserving object semantics.

To capture both semantics and spatial relationships, we jointly model in $f$ two complementary feature fields: one view-invariant $f_{VI}$ and one view-dependent $f_{VD}$, the latter being modeled as a deviation $\delta_{VD}$ of  $f_{VI}$ to ensure consistency between these two:
$$
f_{VD}(X,d)=f_{VI}(X)+\delta_{VD}(X,d)
$$ 
with $X$ the 3D position in the scene and $d$ the direction of observation. For each training sample, the feature $f$ is first decoded into $f_{VD}$ and $f_{VI}$ (each one having its own decoder), which are then decoded into full LLaVA tokens $t_{VD}$ and $t_{VI}$ using the same shared decoder and supervised independently with an MSE against the same ground-truth tokens.

\subsection{Decomposing a NeRF into Objects Hierarchy}
\label{sec:met2}
\textbf{Hierarchical NeRF Feature Field.}
We draw inspiration from Garfield~\cite{kim2024garfield}, which leverages SAM~\cite{kirillov2023segment} masks across views to model implicitly a consistent 3D instance segmentation via a contrastive segment embedding field. In this setting, embeddings associated with rays falling within the same 2D mask are pulled together, while those from different masks are pushed apart, encouraging cross-view consistency. However, unlike Garfield’s continuous spatial scale-space, which encodes scale in terms of spatial volume, we aim to construct a semantic scale-space that reflects discrete hierarchical levels: objects, parts, and sub-parts. This discrete structure aligns better with the goal of structured scene understanding and enables a clear representation of hierarchy. 

We perform this by replacing Garfield’s single scale-conditioned decoder with three separate decoders, all connected to a single set of feature grids and each dedicated to one level of the hierarchy. The training process is adapted accordingly: using the SAM discretization trick introduced in LangSplat~\cite{qin2024langsplat}, we separate the 2D SAM masks into three semantic levels (object, part, subpart). Each set of masks supervises the training of the corresponding decoder and feature field, enforcing a level-specific representation of the scene.

Finally, to both enhance the scene decomposition and enable additional scene understanding downstream tasks (Section \ref{sec:eval_segmentation}), we further augment this field with a CLIP output. For each scale, an additional decoder predicts CLIP features from intermediate segment embeddings, trained using an MSE loss against CLIP embeddings computed from rendered SAM masks. Architecture details can also be found in Section \ref{sec:sup-mat:segmentationField} of supplementary. \\
\textbf{3D Objects Hierarchy Extraction.} 
Now that we have learned a three-level hierarchical feature field, the next step is to construct a true hierarchical scene graph, that is, a tree structure in which any point sampled from the NeRF can be associated with a specific node in the objects hierarchy.

First, to derive full-scene segmentation from our hierarchical feature field, we cluster segment embbedings of a batch of randomly sampled rays using the clustering algorithm HDBScan~\cite{mcinnes2017hdbscan} independently at each semantic scale. To further encourage a scale-specific segmentation, we vary HDBScan's parameters per-scale (see Section \ref{sec:sup-mat:hyperparam} of supplementary). Cluster centroids are computed as confidence-weighted embedding averages, such that any 3D point is assigned to its nearest centroid at each level.

However, this approach is not guaranteed to produce clean hierarchical segmentation, whether in terms of hierarchical misclassifications or reconstruction-based artifacts. 
To improve consistency, we introduce a three-fold refinement step that leverages CLIP features and the multi-scale hierarchy. For each HDBScan's segment, we compute a CLIP centroid and apply the following heuristics across scales: (i) discard noisy segments with high intra-cluster variance or low cardinal; (ii) split under-segmented regions when coarse clusters contain sub-segments with very dissimilar CLIP features, using the finer scale segmentation to compute new centroids; and (iii) merge over-segmented ones if feature centroids (either CLIP or SAM) are nearly identical.

To construct a hierarchical structure from our NeRF-based segmentation, we analyze the relationships between segments in a bottom-up manner. For each finer-level segment, we determine its parent by identifying the coarser-level segment it most frequently co-occurs with, using the HDBScan's rays clustering statistics. This process ensures a consistent and well-structured hierarchy in which each segment is uniquely assigned to a parent, resulting in a coherent and comprehensive decomposition of the scene. 

\subsection{Object-centric Description of the scene}
\label{sec:description}
\textbf{Omnidirectional Visual-Description of a 3D Object.}
For each object $i$, we aim to extract $N_f^i$ features that jointly describe its semantics and spatial relationships. To do so, we randomly cast rays through the supervision images and assign each resulting LLaVA feature to its corresponding object segment. This process continues until each object accumulates $N_f^i$ features. However, to construct an effective object representation, several questions must be addressed regarding feature selection, distribution, and quantity.

First, how can we ensure spatially balanced feature coverage across the whole object ?
To address this, features are allocated regarding the objects hierarchy:
first equally across part regions, then across sub-parts. This guarantees a uniform distribution over the components of the objects. Since the features are permutation-invariant, their internal ordering does not affect downstream processing.

\begin{figure*}[t]
        \includegraphics[width=\textwidth]{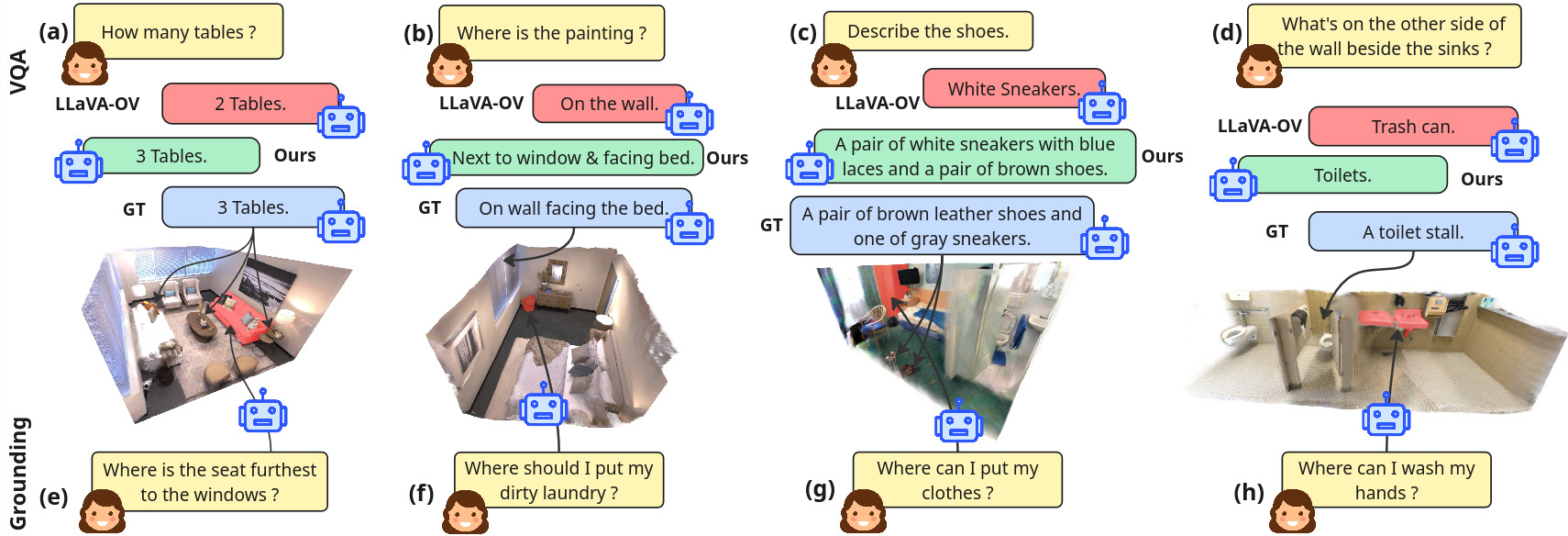}    
    \caption{\textbf{Qualitative performance of our method on 3D VQA and Grounding}. By decomposing into objects the view-dependent features, our method avoids several very common issues in 3D VQA: (a) missing objects due to insufficient sampling, (b) weak inter-object spatial relationships, (c) loss of objects details and (d) bad multi-view understanding (i.e. relations between objects from different images). Our method can also perform grounding from different types of queries ((e),(f),(g),(h)).}
    \label{fig:results}
\end{figure*}

A second key question concerns the type of features that should be selected: should they be purely view-independent (VI), purely view-dependent (VD), or a mixture of both ? 
VI features are better suited for object general semantics, while VD features excel at capturing spatial relationships that are inherently viewpoint-sensitive. 
We thus propose to distribute samples equally between VI and VD features. However, in tasks that involve spatial relationships from an observer’s perspective (e.g. VQA), where references to "left" or "right" are common, we adopt an alternative strategy. We define a canonical viewing direction for each object and extract VD features only from rays that fall within a specified angular threshold of this direction, VI features being extracted otherwise. For simplicity, the canonical direction is defined as the most frequently observed viewing direction of the object, computed via spherical binning over all ray directions used to observe it.

Finally, how many features should be allocated per object ?
We fix the number of features per object to $N_f=W/O$, where $W$ is the maximal VLM context window and $O$ is the number of objects. This strategy ensures uniform object representation. Ablations in the supplementary material confirm that this design leads to improved performance over variable-size allocations (depending on the object complexity for instance). \\ 
\textbf{Object-centric Scene Description.}
The final step consists in aggregating those objects visual descriptions into a coherent scene-level representation, suitable for VLM-based reasoning.
This process also raises important design questions, particularly around how to structure the sequence of object features in the prompt, given the constraints of using a frozen 2D VLM.

Unlike 3D MLLM approaches that introduce positional encodings that require LLM fine-tuning, we explore how the arrangement of objects in the prompt affects how the VLM understands spatial relations across them.

To impose a consistent and layout-aware order, we adopt a radar-inspired sorting strategy. We first compute the 3D centroid of each object, then simulate a sweep originating from the scene center, sorting objects by their polar angle around the vertical axis. To resolve ties between objects with similar angles, we use a secondary term based on their radial distance from the center, with lower weight. This results in a deterministic and geometry-informed object ordering.

Finally, scene description is provided to the VLM as multi-view images, each object being tagged as a virtual image with a unique ID.
This allows us to support grounding tasks: when prompted, the VLM outputs the most relevant ID for a given query, which can be mapped back to the corresponding segment in the NeRF reconstruction.

\section{Experiments}
\label{sec:exps}

\subsection{Implementation Details}
\label{sec:impl}

We implemented our method in the Nerfstudio \cite{tancik2023nerfstudio} framework. Every NeRF is trained with the Nerfacto model, a grid-based NeRF method coupled with several Mip-NeRF-360 \cite{barron2022mip} improvements. Each NeRF trained on ScanNet uses $80\%$ of the images (chosen based on estimated blurriness) for geometry and 400 selected images for the two feature fields. The selection was made by maximizing the scene coverage of the scene, as explained in Section \ref{sec:sup-mat:frameSelec} of the supplementary.
All our experiments (NeRF training and LLM inference) were run on one A100 GPU. 
Additional details on hyper-parameters, evaluation protocols and reproducibility can be found in the supp. mat., as well as additional ablative experiments.

\subsection{3D Visual Question Answering (3D VQA)}
\label{sec:vqa}

\textbf{Datasets and Protocol.} We use ScanNet datasets~\cite{dai2017scannet} for evaluation, specifically the ScanQA~\cite{azuma2022ScanQA} validation set and the MSR3D~\cite{linghu2024msr3d} test set. For ScanQA, we evaluate performance using standard n-gram-based metrics: CIDEr, METEOR, ROUGE, EM@1 and EM@1-Refined. For MSR3D, we adopt their GPT-based correctness score and follow their implementation with GPT-4o as the notation model. \\
\textbf{Qualitative Results.}
Figures~\ref{fig:results} and~\ref{fig:overview} showcase various example with different common errors solved by our method and object-centric VQA (i.e., feeding tokens of specific objects to the VLM rather than the whole scene). \\
\begin{table}[bt!]
\centering
    \setlength\tabcolsep{1pt}
    \setlength\extrarowheight{2pt} 
\resizebox{\columnwidth}{!}{%

    \begin{tabular}{c c ccccc}
     \hline
    Method                            &            Modality               & \multicolumn{1}{c}{CIDEr $\uparrow$} & \multicolumn{1}{c}{METEOR $\uparrow$} & \multicolumn{1}{c}{Rouge $\uparrow$} & \multicolumn{1}{c}{EM@1 $\uparrow$} & EM@1-R $\uparrow$\\ \hline
    \hline
    \textbf{\textit{3D LMMs}}        &           &         &           &           &       &             \\ 
    3D-VisTA* (T)               &  PC       & 69.6    & 13.9      & 35.7      & 22.4  & -     \\
    Chat-3D-v2* (FT)         & PC        & 77.1    & 16.1      &  40.0     & -     & -        \\ 
    LL3DA* (T)               & PC        & 76.8    & 15.9      &  37.3     & -     & -       \\ 
    Scene-LLM*               & PC + I    & 80.0    & 15.8      &  35.9     & -     & -       \\ 
    Scene-LLM* (FT)            & PC + I    & 80.0    & 16.6      &  40.0     & 27.2  & -          \\
    LEO* (T)                 & PC + I    & \textcolor{lightgray}{101.4}   & \textcolor{lightgray}{20.0}      & \textcolor{lightgray}{49.2}      & \textcolor{lightgray}{24.5}  & \textcolor{lightgray}{47.6} \\
    \hline
    \textbf{\textit{2D VLMs} }  &                           &    &       &   &  &       \\ 
    Claude*                   & I        & 57.7    & 10.0      & 29.3      & -     & -       \\ 
    GPT-4V*                   & I        & 59.6    & 13.5      & 33.4      & -     & -           \\
    LLaVA-NeXT-Video*         & I        & 46.2    & 9.10      & 27.8      & -     & -          \\ 
    LLaVA-OV                 & I       & 55.38   & 13.22     & 30.36      & 16.83  & 33.99      \\ 
    LLaVA + Object-based     & I        & 51.78      & 12.87        &29.46  & 16.05     &32.45       \\ 
     \hline
    \textbf{\textit{Hybrid Methods}} &                          &   &       &   &  &       \\ 
    SplatTalk*                 & I       & 61.7    & 14.2    & 32.7      & 17.1    & 32.2  \\
    SplatTalk (FT)*              & I       & \underline{77.5}    & \underline{15.6}    & \underline{38.5}      & \underline{22.4}      & \underline{38.3}   \\
    Ray-NeRF      & I                         & 60.56   & 13.82     &  31.56  & 20.83 & 32.77       \\
    LLaVA$^3$             & I                         &  \textbf{77.69}  &\textbf{ 15.83 }   & \textbf{39.69}  & \textbf{26.00}  &\textbf{ 38.43}       \\ \hline
    \end{tabular}%
    }
    \caption{\textbf{3D VQA Performance on ScanQA validation set.} FT and T refer to fine-tuning and training on ScanQA data. \textcolor{lightgray}{Grey} entries denote  incorporation of ground-truth objects as inputs. "*" refers to results taken from papers and bold indicates best performance among the image-based methods. PC refers to Point Cloud and I to Images.
    }
    \label{tab:vqa-main}
\end{table}
\begin{table}[bt!]
\centering

    \centering
    \setlength\tabcolsep{1pt}
    \setlength\extrarowheight{2pt}
    \resizebox{\columnwidth}{!}{%
    \begin{tabular}{c ccccccc}
    \hline
    Methods & Counting & Existence & Attributes & Spatial & Navigation & Others & Overall\\ \hline
    \textbf{\textit{Baselines}}       &      &      &      &     &       &      &    \\
    LEO*  & 0.8    &   15.5   &  11.8  &  7.3    &  2.3  &    15.3   &    7.8    \\
    
    LLaVA-OV             & 19.20 & 35.00 & 28.48 & 23.36 & 12.50 & 39.94 & 27.87  \\
    Object LLaVA-OV      & 15.19     & 33.95     & 29.18     & 18.85     &  10.79    & 43.31     &  25.02     \\
    SplatTalk*      &  19.6    &    \underline{60.3}  & \underline{44.0}     & \textbf{35.8}    &   \textbf{ 35.5 }  &    \underline{61.8}  & \underline{41.8}   \\
    Ray-NeRF          &  \underline{21.25}    & 52.35     & 36.79     & 30.54     & 15.40     & 53.29     & 30.54      \\
    \hline
    \textbf{\textit{Ablatives}}       &      &      &      &     &       &      &    \\
    + \textit{Object-Based}            &   +4.19   & +14.32     &  +9.31    & -1.86     & +9.22     &  +9.69    &  +6.85     \\ 
    \textit{+ Multi-Scale}         &  +1.35 &  +3.04  &    +3.08  &    +0.65  &    +1.05  &    +3.55  & +2.5       \\ 
    \textit{+ Filtering}             & +0.30   & +0.93     &  +1.42    & +0.36     &  +0.22    & +1.60     & +0.55      \\
    \textit{+ Radar Sweeping}   & +2.05     & +3.28     & -0.11     & +1.48    & +3.21     & +4.18     & +2.09      \\ \hdashline
    $100\%$ VI &  29.14    & 73.92    &  50.49  & 31.17   & 29.1  & 72.31  & 42.53  \\
    $100\%$ VD                 &   29.01 &    71.99  &    49.99  &    31.98  &    30.02  &  73.49  &   42.60   \\
   $50\%$ VI - $50\%$ VD         &   29.34   &    74.13  & 51.02     &  32.43    &  30.20   & 72.99 & 43.84
\\

    \hline
    \textbf{\textit{LLaVA$^3$}}   
             & \textbf{29.51}     & \textbf{75.00}      &   \textbf{51.60}   &   \underline{33.30}   &    \underline{31.35}  &    \textbf{73.99}  &       \textbf{44.89} \\\hline
    \end{tabular}%
    }
    \caption{\textbf{3D VQA performance per question type on MSR3D }test set using their correctness score ($\uparrow$). \textit{Ablatives} first display the relative difference to the previous line down from Ray-NeRF baseline (with $100\%$ VI features) and then uses the full method while only modifying the VI/VD distribution (LLaVA$^3$ uses the Adaptive VI-VD distribution).}
    \label{tab:msr3D}
\end{table}
\textbf{Results on ScanQA.} 
Similarly to SplatTalk evaluation, we compare our methods against several families of baselines on the ScanQA validation set and we report results of each in Table~\ref{tab:vqa-main}. "\textit{3D LMMs}" lists several 3D large multi-modal models trained across a diverse range of tasks. However, some of those models use ScanQA train set, either during their training or their fine-tuning. Please note that \underline{all} those models uses ScanNet and other indoor scenes as training dataset, thus limiting the generalization to other types of scenes and inducing a boosting bias in their performance evaluations.
"\textit{2D VLMs}" refers to models that only process multi-view images inputs. The object-based LLaVA-OV baseline is constructed by applying SAM to multi-view images and extracting LLaVA features from each resulting mask. These per-object features are then fed directly to the LLM. We build this baseline to evaluate the impact of using object-centric inputs from 2D images, in contrast to our proposed 3D object-based pipeline. Lastly, "\textit{Hybrid}" methods only use 2D VLMs but first reconstruct in 3D the scene using NeRF or GS from multi-view images of the scene. Alongside SplatTalk (generalist and LoRA) and our method, we introduce a Ray-NeRF baseline that samples LLaVA visual tokens from random rays across the NeRF field, without object structuring. This baseline isolates the benefit of object-centric feature extraction by contrasting it with an unstructured use of the same NeRF representation.

Most importantly, our base method outperforms all zero-shot NeRF and GS models and 2D VLMs. This demonstrates that we successfully derive a more comprehensive and efficient scene representation for a VLM than traditional multi-view input and random ray sampling. Specifically, both NeRF methods (especially ours) achieve higher performance than LLaVA-OV although we use the same LLM and input feature maps, meaning that although using a NeRF to ensure the use of visual tokens across the whole scene helps ($+5.18$ CIDEr), adding the object-centric representation of LLaVA$^3$ benefits significantly more ($+22.31$ CIDEr). However, the 2D object-based baseline ($-3.60$ against LLaVA-OV) demonstrates that this object-centric decomposition is beneficial only for 3D representation.
When comparing to other paradigms, we notice better or comparable results to most \textit{3D LMMs} (except for LEO, our metrics range between $-4.6\%$ and $+13.2\%$ of the other baselines with an average of $+2\%$ across all metrics), despite not using any 3D-specific information, nor training our VLM on ScanNet data. \\
\textbf{Results on MSR3D.} To evaluate the impact of our different contributions, we showcase quantitative results in Table~\ref{tab:msr3D} on the MSR3D dataset, which decomposes into multiple question types, allowing fine-grained understanding of what brings each contribution. In addition to the LLaVA-2D baselines, SplatTalk and LEO, we progressively build LLaVA$^3$ on the Ray-NeRF baseline by adding our contributions one by one to demonstrate their individual impact.

First, we analyze our contributions. Decomposing the scene into objects has the highest effect on overall performance ($+6.85$), as it ensures a balanced distribution of the sampling across the scene. More in-depth balancing of the features via hierarchical sampling further increases the results ($+2.5$); especially for questions related to visual appearance (e.g. existence and attributes with resp. $+3.04$ and $+3.08$), as it gives the model more exhaustive information on each object. As expected, improving the scene decomposition quality by adding filtering steps slightly improves the results globally. 
Ordering the scene description via our radar sweeping strategy also noticeably helps the VLM, with a $2.09$ score bump.
Regarding the view dependency, although replacing VI features with VD features results in similar overall performance (but still different detailed distribution), combining both features in an even split improves the performance. It appears that VI features help the model to reason over descriptive queries (counting, existence, attributes), while the VD features help with spatial and navigation questions. Finally, the adaptive VI-VD used in LLaVA$^3$ is better than splitting evenly the features, as choosing one canonical view direction helps follow the observer's perspective. 

Our full model outperforms most comparative baselines across the majority of categories. LEO, which is specifically trained for ScanQA-style queries, struggles to generalize to this dataset. The 2D baselines show overall lower performance, largely due to the small number of input images. Similarly, the other hybrid methods such as SplatTalk and our Ray-NeRF baseline under-perform on most categories, with respective drops of $-3.09$ and $-14.35$ points.

\subsection{Grounding}
\label{sec:grounding}

\textbf{Datasets.} We evaluate grounding on ScanNet-based Sr3D+ and Nr3D, which provide natural language queries with ground-truth object instance IDs. We use a subset of ScanNet scenes: 0011\_00, 0030\_00, 0046\_00,0086\_00, 0222\_00, 0378\_00, 0389\_00 and 0435\_00. \\
\textbf{Protocol.} 
Following standard baselines, we perform grounding by retrieving a binary segmentation point cloud from our segmented NeRF using the instance ID predicted by the LLM. To convert our NeRF into a point cloud, we adopt the OpenNeRF protocol: for each supervision view, we render the segmentation mask and back-project it onto the point cloud. For each point, we average the instance class feature across views and use the centroids to determine the segmentation.  We then compute standard detection-style metrics: a prediction is correct if the segmentation IoU exceeds a threshold. We report accuracy at 0.1 and 0.25 IoUs, following prior works. \\
\textbf{Results.} 
Results are reported in Table~\ref{tab:grounding}, with qualitative examples provided in Figure~\ref{fig:results}. First, our method heavily outperforms existing NeRF-based open-vocabulary baselines such as LeRF and OpenNeRF (resp. $+245\%$ and $+60\%$ in average). This improvement is expected, as those methods rely solely on CLIP-based similarity measures, which are effective when object names are explicitly mentioned in the query. In contrast, grounding queries are phrased as spatial indications, making such non-reasoning approaches less effective. The ConceptGraph baseline, which stores an explicit caption per-object, achieves lower performance compared to ours, highlighting the fact that our object-centric tokenization gives the LLM the ability to reason over objects without any retraining. Using our hierarchical representation, we can also perform sub-scale grounding, as illustrated in Figure~\ref{fig:overview}.

\begin{table}[tb!]
\centering
    \setlength\tabcolsep{1pt}
    \setlength\extrarowheight{2pt} 
\resizebox{0.8\columnwidth}{!}{%
\begin{tabular}{c cc cc}
\hline
\multirow{2}{*}{Methods} & \multicolumn{2}{c}{Sr3D++}                                   & \multicolumn{2}{c}{Nr3D}                                   \\ \cline{2-5}
         & \multicolumn{1}{c}{Acc@0.1 $\uparrow$} & \multicolumn{1}{c}{Acc@0.25 $\uparrow$} & \multicolumn{1}{c}{Acc@0.1 $\uparrow$} & \multicolumn{1}{c}{Acc@0.25 $\uparrow$} \\ \hline
LeRF                     &  6.88  & 1.97  &  5.53     &  1.43    \\ 
OpenNeRF                 & 9.07   &    4.03     & 9.70         &  5.11    \\  \hline
ConceptGraph             &   \underline{13.3}  &  \underline{6.2} &  \underline{16.0}    &    \underline{7.2}  \\ \hline
LLaVA$^3$                    & \textbf{14.41}  &\textbf{ 6.54}     &  \textbf{16.17}   &   \textbf{7.81}   \\
\hline
\end{tabular}%
}
\caption{\textbf{3D Grounding Performance} on Sr3D+ and Nr3D.}
\label{tab:grounding}
\end{table}

\subsection{Semantic Segmentation and Other Segmentation Tasks}
\label{sec:eval_segmentation}

By leveraging the SAM-CLIP feature field (Section~\ref{sec:met2}), we enable a broad range of segmentation tasks, a fundamental aspect of downstream 3D scene understanding. LLaVA$^3$ supports semantic segmentation (see Figure~\ref{fig:overview}), but also  open-vocabulary segmentation and instance segmentation, all within a hierarchical framework.

For semantic segmentation specifically, evaluation on standard benchmarks using both Replica and ScanNet scenes showcases that our method significantly outperforms other NeRF-based approaches such as LeRF~\cite{kerr2023lerf}, OpenNeRF~\cite{engelmann2024opennerf}, and DiSCO-3D~\cite{petit2025disco3d}, achieving respectively up to +146\% mIoU / +91\% mAcc, +21\% / +7\%, and +15\% / +14\% gains. It can also be noted that it compares favorably with explicit scene-graph-based models, outperforming ConceptGraph (+36\% mIoU and +20\% mAcc) and matching HOV-SG (+4\% and -3\%). Additional results, figures, and analyses are provided in the supplementary.

\section{Conclusion}
\label{sec:ccl}

We presented LLaVA$^3$, a novel object-centric approach that improves the 3D scene understanding ability of VLMs from multi-view images without fine-tuning. Inspired by the principles of Cubism, our method decomposes the scene into objects hierarchy and compute for each of them an omnidirectional 2D visual representation that captures both its semantic and  spatial relationships with the scene. Experimental results demonstrate that it enables VLMs to reason more effectively over 3D content, avoiding common pitfalls such as object duplication or limited context windows, and outperforming other VLM-based solutions. LLaVA$^3$ enables a large variety of downstream tasks, including 3D VQA, 3D grounding and 3D semantic segmentation.

\section*{Acknowledgements} This publication was made possible by the use of the CEA List FactoryIA supercomputer, financially supported by the Ile-de-France Regional Council.

\bibliography{aaai2026}

\clearpage




\begin{center}
  \LARGE \textbf{Supplementary Material} \\[0.5em]
  
\end{center}

\section{Method}

\subsection{Optimal Frame Selection.}
\label{sec:sup-mat:frameSelec}
Video sequences of large environments typically contain many redundant frames, making it computationally expensive (and often unnecessary) to encode every image with LLaVA and SAM. To balance efficiency and feature diversity, we sub-sample the training images used for learning the feature fields (while still using all frames for geometry). Rather than sampling uniformly, we apply a greedy selection strategy that maximizes coverage in 3D camera space. At each step, we select the frame most dissimilar from the current set, based on camera position and orientation. This leads to a diverse and efficient subset of frames for distillation. We evaluate in Section \ref{sec:sup-mat:ablative} the impact of this algorithm.

\subsection{Details on the LLaVA Feature Field}
\label{sec:sup-mat:LlavaField}
\begin{figure}[b]
    \includegraphics[width=\columnwidth]{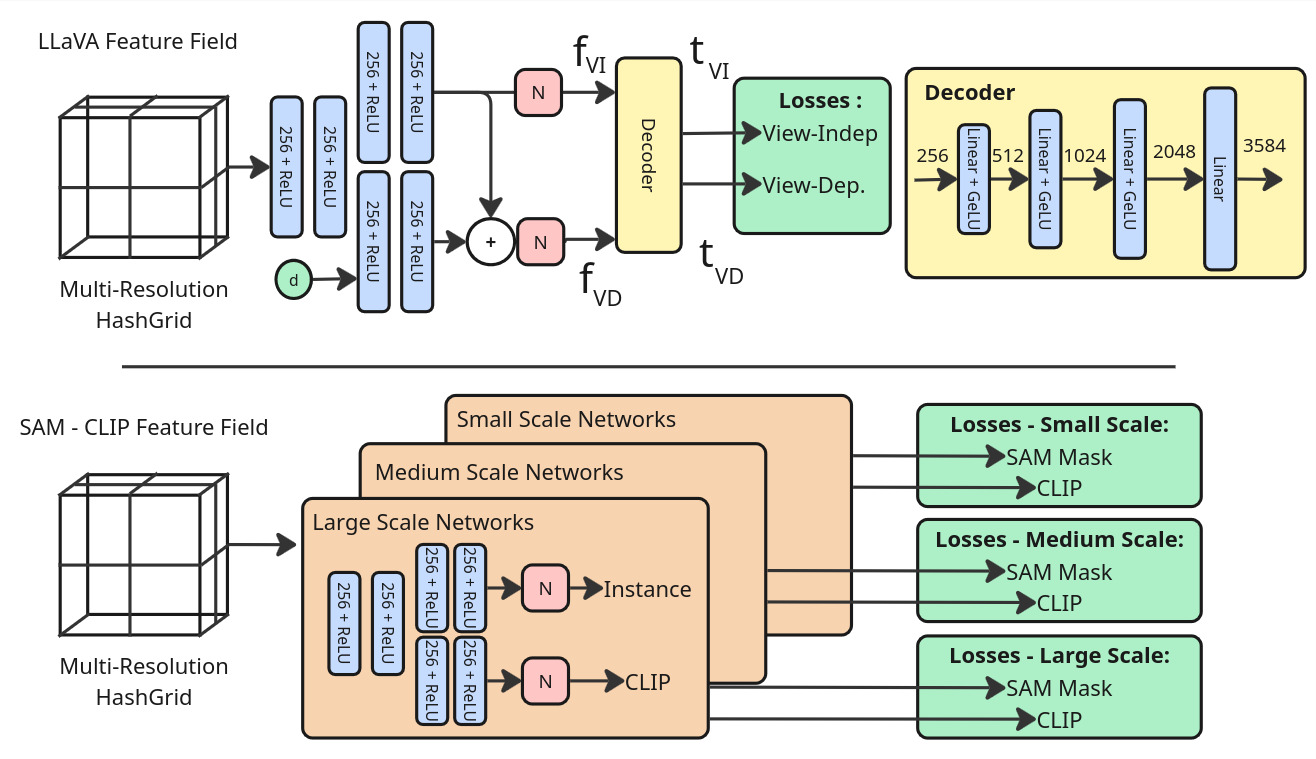}    
    \caption{Precise Architecture of the Feature Fields. $N$ stands for "Normalization".}
    \label{fig:sup-archi}
\end{figure}

Figure~\ref{fig:sup-archi} illustrates the exact network architectures used for reconstructing the feature fields. For the LLaVA feature field, we detail both the decoder architecture (largely inspired by SplatTalk)and the view-dependency module. 

We supervise the LLaVA feature field using a standard Mean Squared Error (MSE) loss on the predicted features. Given that LLaVA tokens are high-dimensional and unnormalized, we also experimented with a Relative MSE (RMSE) loss, which adjusts each feature dimension's contribution based on its magnitude. Intuitively, this choice aimed to make the loss invariant to absolute feature scale, which one might expect to be irrelevant for downstream performance. However, in practice, the RMSE-trained field consistently led to degraded performance. Regardless of the sampling strategy, LLaVA tokens extracted from this field yielded severe hallucinations when fed into the VLM, indicating that the scale of individual dimensions plays a crucial role in the interpretability of the features by the VLM. As a result, we retain the use of standard MSE loss for all experiments.

\subsection{Details on the Segmentation Feature Field}
\label{sec:sup-mat:segmentationField}
The architecture used for the SAM-CLIP feature field is also provided in Figure~\ref{fig:sup-archi}, though its design choices are relatively straightforward and primarily guided by implementation simplicity. Additionally, we describe here the training process used for this field. We follow Garfield idea where we apply a contrastive loss based on the SAM masks. During optimization, we sample rays randomly across the supervision images and sample randomly a mask for each scale across all masks overlapping the specific pixel associated to each ray. Those rays are then sorted by images and we create pairs of rays within each image. Their associated features are pulled closer if their corresponding 2D masks match, and pushed apart otherwise. We apply this process once per scale independently from one another. For each scale, we thus compute per-image the following losses and sums them:
\begin{equation}
\mathcal{L}_{pull} = ||e_A - e_B || 
\end{equation}
\begin{equation}
\mathcal{L}_{push} = ReLU(m - ||e_A-e_B||)
\end{equation}
where m is the fixed lower bound distance, or margin.


During training of the LLaVA feature field using the Garfield framework, we observed numerical instabilities, particularly when using mixed-precision optimization. These issues manifested as infinite values during loss computation, typically caused by large intermediate values in the computation graph. To address this, we restructured the loss function to improve numerical stability by applying logarithmic operations where appropriate. This adjustment effectively mitigated overflow issues and allowed stable convergence under mixed-precision settings.

\subsection{Discussion on the choice of those two fields}
An important but non-mentioned choice of our method is the complete independence between our two feature fields. LLaVA field is learnt from encoded feature maps from the whole images and does not use the segmentation field in any way and vice-versa. However, it would have been possible to interlace both training processes to maybe improve the performances. One main idea would be to follow our implementation of CLIP per-mask encoding in the LLaVA field and compute a per-mask LLaVA feature map for LLaVA supervision. However, we chose not to pursue this direction for the following reasons:
\begin{itemize}
    \item \textbf{Computational Intractability.} Although the precise number varies, SAM model easily produces more than 100 mask per image across all scales. Computing and storing feature maps of dimension $400\times100\times27\times27\times3584 (\approx105*10^9)$ obviously results in impossibly long processing time and most importantly $>100$Gb per-scene checkpoint, which is not really realistic for real-life applications.
    \item \textbf{Loss of Contextual Information.} Mask-level supervision for LLaVA features would eliminate spatial context by encoding each object in isolation. This absence of surrounding information would severely hinder the model’s capacity for spatial reasoning and scene-level understanding.
    \item \textbf{Modularity and Flexibility.} Maintaining separate feature fields enables modular use of each component depending on the target application. For instance, tasks that do not require a language model (e.g., semantic or instance segmentation) can be performed solely using the segmentation field, reducing both memory footprint and training complexity.
\end{itemize}





\subsection{Hyperparameters}
\label{sec:sup-mat:hyperparam}

Our method is composed of several key components, each governed by its own set of hyperparameters. However, it is important to emphasize that these hyperparameters are not critical to the overall functionality of the approach. They primarily serve to balance trade-offs between performance, memory consumption, and computation time. The core behavior and effectiveness of the method remain consistent across a broad range of reasonable values. Below, we provide the specific hyperparameter settings used in the experiments reported in the paper.

\begin{figure*}[tb!]
\centering
    \includegraphics[width=0.8\textwidth]{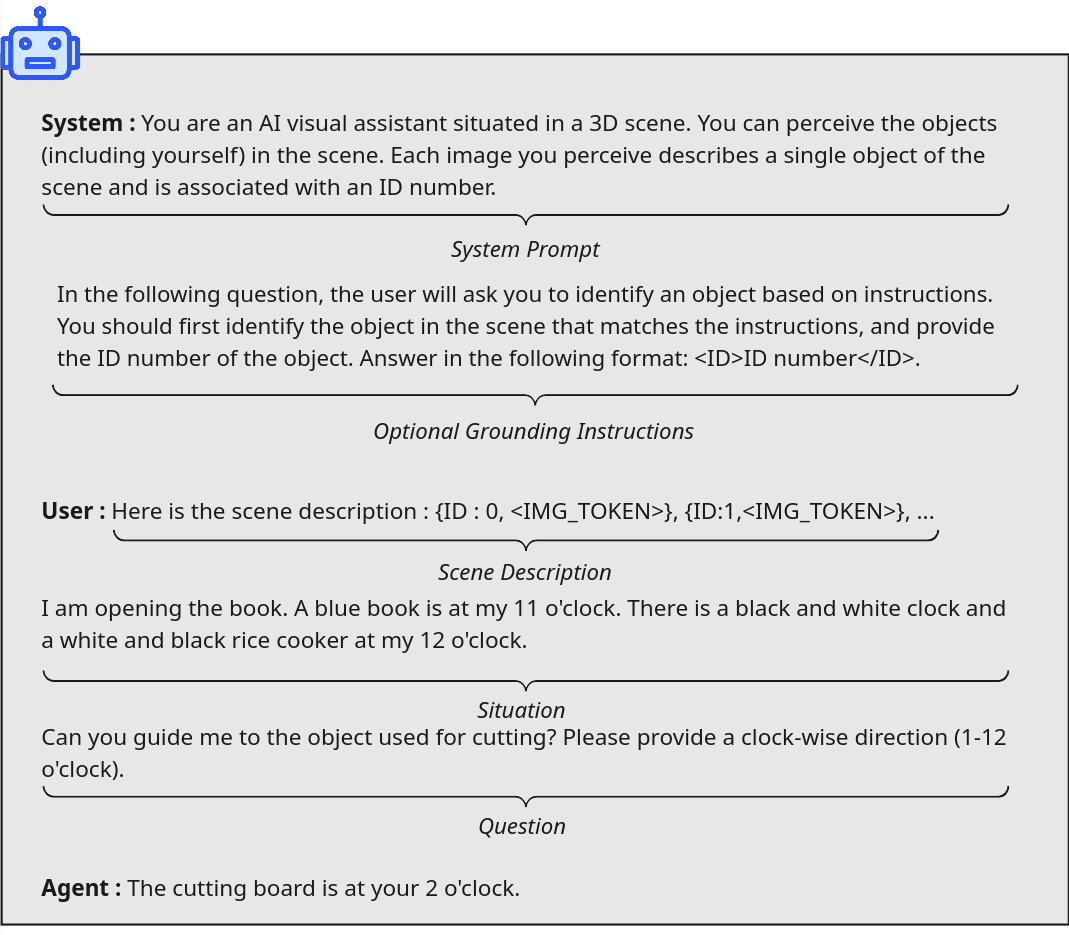}    
    \caption{Prompt Example in our model for a MSR3D example question.}
    \label{fig:sup-prompt}
\end{figure*}

\begin{itemize}
    \item \textbf{ScanNet Reconstruction.} ScanNet data often includes frames affected by motion blur, which can degrade the quality of 3D reconstruction and downstream processing. To mitigate this, we apply a Laplacian variance filter to quantify sharpness and discard the 20\% most blurred images in each scene. Additionally, the undistortion step in the ScanNet preprocessing pipeline introduces black border artifacts. To address this, we crop 20 pixels from each edge of every image, ensuring clean inputs to the reconstruction process. As for NeRF training, we do not require fixed hyperparameters beyond ensuring a visually and geometrically accurate reconstruction. Note that we use both camera pose optimization and per-image appearance embeddings to enhance rendering fidelity.

    \item \textbf{Feature Field Reconstruction.} We follow the previously described architecture and have two different feature fields with different hyperparameter sets each. Regarding the segmentation field, we use standard parameters, with 16 levels of grids with resolutions ranging from 16 to 2048. The constrastive SAM feature size (outputted from the SAM MLPs) is of 64 for each scale, which is smaller than what's used in Garfield, as the use of three discrete levels requires less complex features than the continuous scale groupings, but most feature sizes are usable. Because, the LLaVA feature field needs to store complex features, we take inspiration from LeRF configurations and use 32 layers of hashgrids with resolutions from 16 to 512. For both fields, each grid has a feature size of 8.

    \item \textbf{Segmentation.} We segment the scene by applying the HDBScan algorithm on each individual scale. We draw $3*8192$ rays and use them for all three HDBScan with each different hyperparameters to encourage coarser clusters as the scale increases. Specifically, for the small, medium, and large scales, we set empirically the minimum cluster size to 10, 30, and 50 respectively; the minimum samples to 3, 5, and 10; and the cluster selection epsilon to 0.01, 0.05, and 0.1.

    \item \textbf{Filtering and Feature Extraction.} Our filtering hyperparameters have been tuned empirically from a few test scenes and then fixed for every experiments. Regarding the noisy classes suppression, we set a minimum of 10 samples per class and an instance and CLIP variances percentiles of 85\%. The over-segmentation merging and under-segmentation splitting uses respectively similarity thresholds of 0.85 and 0.75. Regarding the feature extraction, we extract a total of $30 000$ rays distributed amongst the objects, which is slightly less than the maximum context window size. The following ablative experiment demonstrate the increase of performance.

    \item \textbf{LLM and Image Encoders.} We use in all our experiments \textit{siglip-so400m-patch14-384} as LLaVA's visual encoder and \textit{qwen2-7b} as LLaVA's LLM, from LLaVA-OV pre-trained models. Regarding the SAM-CLIP feature field, we use the SAM from LangSplat codebase (i.e. ViT-H) and CLIP ViT-H-14 as used in ConceptGraph and HOV-SG experimentations.
    
\item \textbf{Prompt.} We provide the full prompt used in our model in Figure~\ref{fig:sup-prompt}. That said, we observed that even a much simpler variant where the LLaVA image features are directly replaced by our per-object features without explicitly informing the model that each “image” represents an object from the same 3D scene produces kind of similar results.
\end{itemize}


\subsection{Compute Ressources.}

All experiments were conducted on a single NVIDIA A100 GPU.. The same GPU was used for all stages, including NeRF optimization, feature extraction, and VLM inference.

\paragraph{Hardware and Memory Considerations} The most memory-intensive component is the Vision-Language Model (VLM) inference, which requires enough VRAM to fit the chosen model. During NeRF optimization and feature extraction, memory consumption can be flexibly managed by adjusting the number of sampled rays per batch and the number of input images. Reducing these parameters allows the method to run on smaller GPUs at the cost of increased training time.

\paragraph{Runtime Breakdown} Our pipeline involves several stages that are each executed once per scene:
\begin{itemize}
    \item SAM-based Preprocessing: Each image undergoes segmentation preprocessing, which takes approximately 8 s per image. In our standard experiments (approx.200 images per scene), this step therefore takes around 25 min in total, but the number of processed images can be substantially reduced without significantly affecting performance (see Supplementary Sec.~\ref{sec:sup-mat:ablative}).
    \item NeRF Optimization: The joint training of radiance, object, and LLaVA-conditioned fields takes roughly 15–20 min per scene to reach a qualitatively stable reconstruction.
    \item Feature Extraction: The extraction of semantic and object features is lightweight, requiring only a few seconds to one minute per scene.
\end{itemize}

During the VLM inference, for each textual query, the VLM generates an answer at its usual language model speed, typically ~5 s per question on our setup.

Overall, the NeRF optimization phase constitutes the majority of the compute time, while the remaining steps are relatively fast. All timings are reported for a single scene and can be scaled proportionally depending on the dataset size and number of questions.

\subsection{Limitations}
An important limitation of our approach lies in its per-scene processing requirement: for each new environment, a NeRF must be trained and its associated feature fields (SAM and LLaVA) computed. While this introduces a one-time computational overhead, especially during the preprocessing of image features, it avoids the need for large-scale retraining or scene-specific finetuning typically required by feed-forward methods such as SplatTalk or 3D MLLMs. As a result, our method remains scene-agnostic by design, enabling zero-shot generalization to diverse environments, including out-of-distribution or non-indoor scenes, without additional data collection or supervision. This trade-off, while involving per-scene effort, ensures strong adaptability and broader applicability in real-world settings.

\section{Experiments}

\subsection{Semantic Segmentation}

As stated in Section~\ref{sec:eval_segmentation}, we evaluate our SAM-CLIP feature field on the semantic segmentation downstream task.

\begin{figure}[tb!]
    \includegraphics[width=\columnwidth]{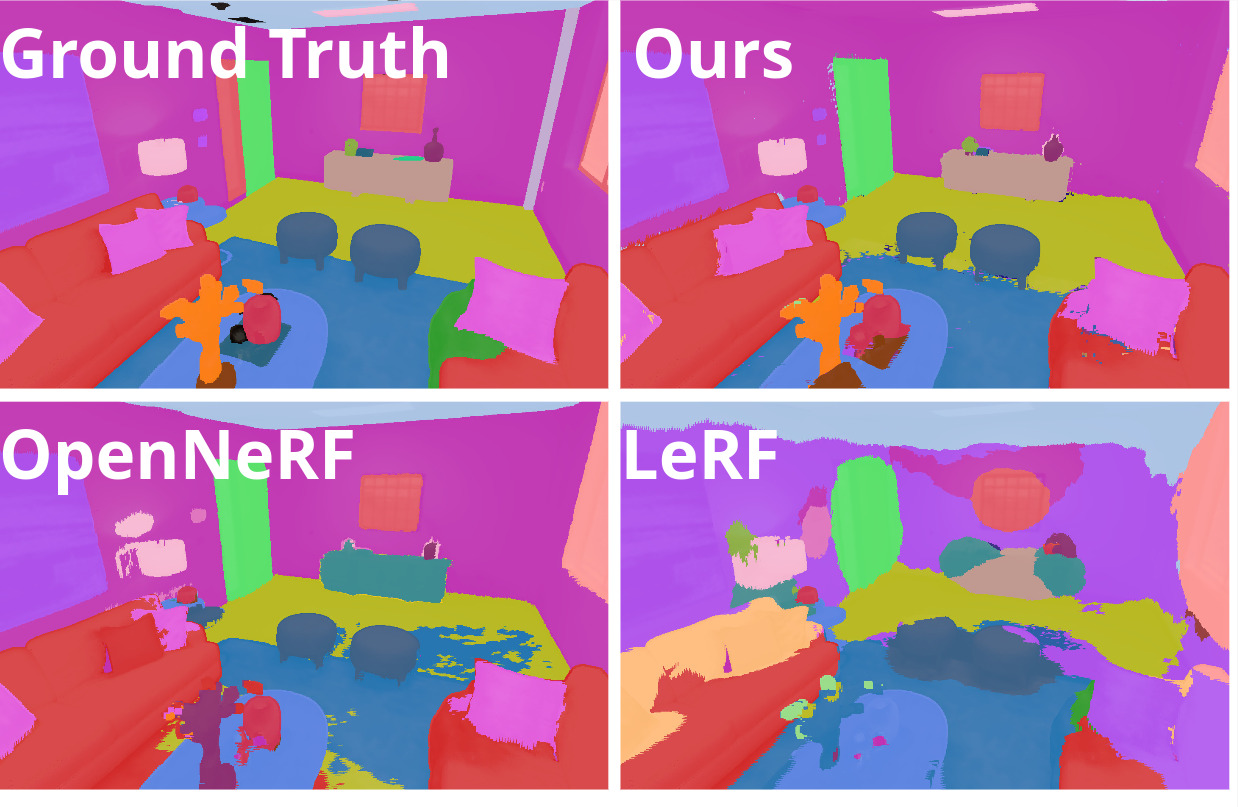}    
    \caption{NeRF Semantic Segmentation Performance on Replica's \textit{room0} scene.}
    \label{fig:sup-semseg}
\end{figure}
\paragraph{}

\paragraph{Protocol.}
To assign semantic labels to each 3D point, we compute per-point and per-scale CLIP embeddings from the NeRF and compare those to the encoded text labels. For mono-scale ablations, predictions are made using features from a single scale while the multi-scale version assigns to each point the label with the highest similarity across all scales. We report mean Accuracy and mean IoU.  \\
\textbf{Results.}
Amongst NeRF-based methods, ours stands out by achieving significantly better results by combining SAM’s high-quality masks with CLIP’s expressive features. Unlike LeRF which relies on DINO regularization and OpenNeRF/DiSCO which leverage less expressive OpenSeg models, our method benefits from an accurate and geometry-aligned supervision signal, enabling more reliable and detailed scene understanding.
Compared to graph-based methods, we achieve better results on most metrics and datasets, showing that continuous NeRF fields can yield high-quality segmentations without relying on explicit point clouds representations.

Ablations confirm the benefit of multi-scale decomposition, with larger scales improving performance due to more coherent object-level masks.
Figure \ref{fig:overview} shows a snippet of semantic segmentation capacities and our supplementary showcases other examples.

Our method outperforms prior NeRF-based approaches by combining SAM’s high-quality masks with CLIP’s expressive features. Unlike LeRF (DINO) or OpenNeRF/DiSCO (OpenSeg), our supervision is both accurate and geometry-aligned, enabling more reliable scene understanding. Compared to graph-based HOV-SG, we achieve better results on most metrics and datasets, showing that continuous NeRF fields can yield high-quality segmentations without relying on explicit 3D graphs.

Ablations confirm the benefit of multi-scale decomposition, with larger scales improving performance due to more coherent object-level masks. 

Figure~\ref{fig:sup-semseg} shows semantic segmentation comparisons on one scene of Replica against other NeRF baselines.

\begin{table}[tb!]
\centering
\resizebox{0.8\columnwidth}{!}{%
\begin{tabular}{cc cc cc}
\hline
\multirow{2}{*}{Methods}                  & \multicolumn{2}{c}{ScanNet}     & \multicolumn{2}{c}{Replica}     \\ 
          & \multicolumn{1}{c}{mIoU $\uparrow$} & mAcc $\uparrow$ & \multicolumn{1}{c}{mIoU $\uparrow$} & mAcc $\uparrow$ \\ \hline
LeRF              & \multicolumn{1}{c}{8.93}     &  15.77    & 10.49    &  22.02    \\ 
OpenNeRF          & \multicolumn{1}{c}{18.63}     &  33.34    & \multicolumn{1}{c}{19.08}      &  31.96    \\ 
DiSCO             & \multicolumn{1}{c}{20.61}     &  30.75    & \multicolumn{1}{c}{20.76}      & 30.19      \\
 \hline 
ConceptGraph      & \multicolumn{1}{c}{16.42}     &   27.60   & \multicolumn{1}{c}{18.72}      & \underline{30.54}     \\ 
HOV-SG   & \underline{22.43}     &  \textbf{43.81}    & \underline{23.16}     &  30.40    \\ \hline
\textit{Ours - Small  }   & \multicolumn{1}{c}{ 17.59}     &  27.22   & \multicolumn{1}{c}{17.45}     &   21.47   \\ 
\textit{Ours - Medium }    & \multicolumn{1}{c}{21.94}     &  32.92    & \multicolumn{1}{c}{22.06}     &   27.85   \\ 
\textit{Ours - Large }   & \multicolumn{1}{c}{22.76}     & 34.26    & \multicolumn{1}{c}{22.96}     &   29.67   \\ 
Ours - All Scales & \multicolumn{1}{c}{\textbf{23.16}}     &  \underline{35.92}    & \multicolumn{1}{c}{\textbf{24.35}}     &   \textbf{33.60}   \\
\hline
\end{tabular}%
}
\caption{3D Semantic Segmentation Performance.}
\label{tab:semseg}
\end{table}

\subsection{Ablative Experiments}
\label{sec:sup-mat:ablative}

In the following, we present additional ablation experiments and extend those introduced in the main paper to further assess the impact of various components and design choices in our method. Unlike the previous ablations which were conducted on the test set of MSR3D, these additional evaluations are performed on a subset of four ScanNet scenes (scene0000\_00, scene0030\_00, scene0046\_00, and scene0086\_00), using the ScanQA dataset for the Visual Question Answering (VQA) task. We report and discuss the results accordingly. \\
\paragraph{Impact of Number of Views and View Selection.}
To reduce pre-processing time, our main paper proposes sub-sampling the number of images used for SAM and LLaVA encodings, from thousands down to just 400. We employ a Coverage Maximization strategy (detailed in Section \ref{sec:sup-mat:frameSelec}) to select the most informative views. In Table \ref{tab:vqa-view-selection}, we compare this strategy against two other image selection methodologies (random image sampling and uniform linear sampling) across three paradigms: our method, LLaVA-OV, and a NeRF baseline with random ray-based feature sampling. We also evaluate performance across varying numbers of selected images.

As expected, Coverage Maximization consistently outperforms other sampling strategies, particularly when the number of views is limited. Increasing the number of images generally improves performance, although gains taper off beyond a certain point. Interestingly, we observe that the 2D baseline LLaVA-OV performs better than NeRF-based methods when using very few views (i.e. 10), which aligns with expectations: NeRF requires dense coverage to reconstruct coherent feature fields (both LLaVA and segmentation), whereas LLaVA-2D is designed to handle sparse multi-view inputs. However, as the number of views increases, NeRF-based methods surpass the 2D baseline. Especially, our LLaVA$^3$ outperforms the other two baselines, as demonstrated in the main paper.

Note that LLaVA-2D is constrained by a fixed context window and can only accept up to 44 images at once.

\begin{table*}[tb!]
\resizebox{\textwidth}{!}{%
\begin{tabular}{c c cc cc cc cc cc}
\hline
                                   &    Num Images               & \multicolumn{2}{c}{10}  &
                                   \multicolumn{2}{c}{44}  & \multicolumn{2}{c}{100} & \multicolumn{2}{c}{200} & \multicolumn{2}{c}{400} \\ 
Paradigm                           & Sampling Method   & CiDER & EM@1-R    &  CiDER & EM@1-R    &        CiDER & EM@1-R    & CiDER & EM@1-R  & CiDER & EM@1-R   \\ \hline
\multirow{3}{*}{LLaVA-2D}          & Random           &  78.27      &  38.37    &  81.29   &  40.65 & -   & -  & -  & -  & - & - \\
                                   & Linear            &   80.67     &  39.19    &    81.09    & 39.96    &  -  &  - & -  & -  & - & -  \\ 
                                   & Maximize Coverage &  82.05     &  40.14    &  83.30    &  41.24    &  -  & -  &  - & - & - & - \\ \hline
\multirow{3}{*}{Random Ray Sampling NeRF}              & Random            &     66.29  &  31.60   &  70.73     &  33.26    & 74.06  & 34.58  & 80.79  & 37.21 & 81.15 & 38.43 \\ 
                                   & Linear            &  68.61      &  33.51    &  74.93     & 36.41    &  77.38  &  37.33 & 81.66  &    38.31 & 81.42  & 38.44 \\
                                   & Maximize Coverage &   69.87     &  34.61    &  74.86     & 36.54    & 78.86   & 37.34  & 33.48  & 35.96  & 81.85 & 36.79 \\ \hline
\multirow{3}{*}{LLaVA$^3$} & Random            &    72.35    &  33.84    &  85.41    &  39.06   & 87.79   & 40.88  & 91.61  & 43.33 & \underline{96.64} & \underline{46.06} \\
                                   & Linear            &  76.93      & 38.19     &       84.99 &    40.12 &  87.65  & 41.04  & 93.95  &  45.38 & 96.43 & \underline{46.06} \\ 
                                   & Maximize Coverage &   79.29     &  40.06    & 85.42       & 42.00    & 89.69   & 43.20  & 93.37  & 45.32   & \textbf{97.82} & \textbf{46.24} \\ \hline
\end{tabular}%
}
\caption{Performances on VQA (ScanQA subset) for different number of encoded images and View Selection algorithms. The experiments are done on the 2D baseline, the NeRF random sampling baseline and our method.}
\label{tab:vqa-view-selection}
\end{table*}

\paragraph{Compressing LLaVA Tokens in Feature Field.}
One key characteristic of LLaVA visual tokens is their high complexity: they are unnormalized, sparse, and extremely high-dimensional (3584 dimensions). As a result, training a feature field to regress these tokens requires stabilizing the learning process through the introduction of an intermediate, lower-dimensional representation. As explained in Section~\ref{sec:met1}, unlike SplatTalk which employs a pre-trained, per-scene auto-encoder to learn stable features directly inside the field, we opt to jointly learn both a stable feature field and a decoder that reconstructs the full LLaVA tokens.

In Table~\ref{tab:vqa-ae}, we compare several strategies for compressing LLaVA features: (i) using a frozen auto-encoder trained prior to the field (i.e. as in SplatTalk), (ii) jointly training the field and decoder from scratch during scene optimization, and (iii) a hybrid strategy where we first pre-train the auto-encoder slightly and then fine-tune the decoder jointly with the field. We report both VQA performance and the average training loss of the feature field to highlight the limitations of using a frozen auto-encoder, which introduces a “double compression” that hampers performance.

As expected, we observe improved performance when the decoder is allowed to update during feature field training, effectively avoiding the drawbacks of double compression by enabling end-to-end optimization. This benefit is reflected in both the final VQA scores and the reduced average training losses. Interestingly, results are similar whether or not the decoder is pre-trained prior to joint training, suggesting that the pre-training step is not strictly necessary. A more detailed analysis reveals that a pre-trained decoder leads to faster convergence of the LLaVA feature field, though the overall impact on final performance remains limited. 

\begin{table}[tb!]
\resizebox{\columnwidth}{!}{%
\begin{tabular}{c c ccc}
\hline

Pre-Trained AE   & Joint Training of Decoder   & CiDER & EM@1-R    &      Training Loss \\ \hline
 \ding{51}       & \ding{55}            &   79.63    &  39.88    &   6.01  \\
  \ding{51}     & \ding{51}            &   97.82     &  46.24   &    3.83   \\ 
\ding{55}       & \ding{51}      &    96.77   & 46.22     &       3.85  \\ \hline

\end{tabular}%
}
\caption{Performances on VQA for different methods of tokens compression.}
\label{tab:vqa-ae}
\end{table}

\paragraph{Impact of View Direction Dependent Features.}
We evaluate the effect of incorporating view dependency into the LLaVA features for the VQA task by testing different sampling strategies per object, meaning that we vary the proportion of view-independent (VI) and view-dependent (VD) features. Results are reported in Table~\ref{tab:vqa-vd}.

We begin by assessing whether explicitly supervising both VI and VD channels is necessary. Results indicate that without this dual supervision, performance drops significantly. This suggests that the VD-based deformation alone is insufficient for the network to naturally learn meaningful VI features.

Next, we activate the dual loss and vary the proportion of VI and VD features. We find that combining both feature types consistently outperforms using only one, highlighting their complementary roles. Finally, our proposed adaptive VI-VD selection strategy yields the best results. This is expected, as it not only leverages both feature types but also constrains VD features to a dominant direction, which helps the LLM better reason about spatial relationships. These results are consistent with those from MSR3D, described in Section~\ref{sec:vqa}, where we use the question categories decomposition to further analyze the impact of combining VI and VD features.

\begin{table}[tb!]
\resizebox{\columnwidth}{!}{%
\begin{tabular}{c c c c cc}
\hline

VI \%  & VD \%   & Adaptive VI-VD & VI-VD Double Loss & CiDER & EM@1-R   \\ \hline
  100       & 0            &   \ding{55}    & \ding{55}  &  72.73    & 31.62     \\
  100       & 0            &   \ding{55}    & \ding{51}  &    90.82  &   42.64   \\

 0      & 100            &   \ding{55}   & \ding{51}   &    88.40  &  42.33    \\
  50     & 50            &     \ding{55}  & \ding{51}   &   94.12   & 43.20     \\ 
-       & -   &    \ding{51}  & \ding{51}   &    97.82 & 46.24            \\ \hline

\end{tabular}%
}
\caption{Performances on VQA (ScanQA subset) with different View Direction dependent features configuration.}
\label{tab:vqa-vd}
\end{table}

\begin{figure}[tb!]
    \includegraphics[width=\columnwidth]{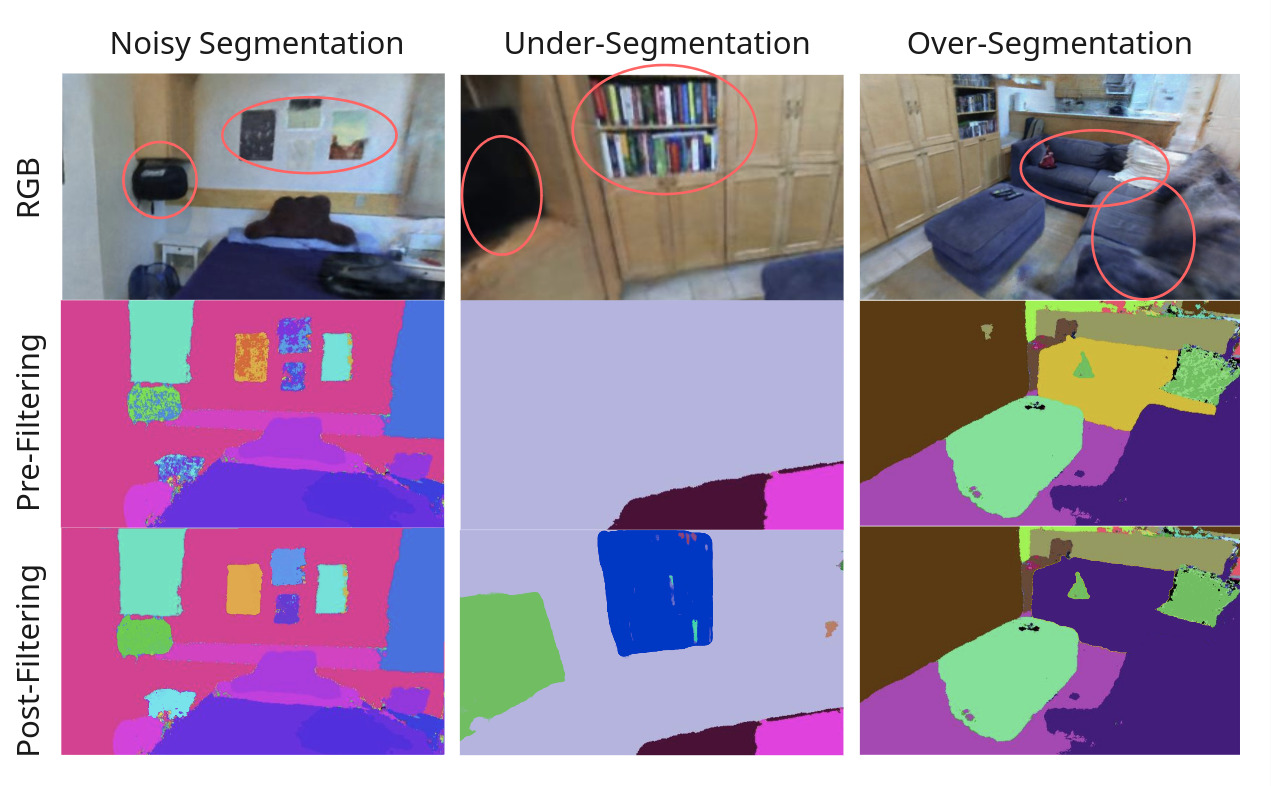}    
    \caption{Examples of the Impact of Filtering on our segmentation.}
    \label{fig:sup-filtering}
\end{figure}

\paragraph{Filtering Impact.}
As described in Section~\ref{sec:met2}, applying HDBScan to the segmentation features can introduce segmentation errors that negatively affect downstream performance when features are later extracted and fed to the LLM.

We identified three primary types of errors: over-segmentation, under-segmentation, and noisy segmentation. Figure \ref{fig:sup-filtering} provides qualitative examples of each issue, both before and after filtering, to illustrate the necessity of the filtering step.

Table \ref{tab:vqa-filtering} presents the quantitative impact of each filtering sub-step on the ScanQA scene subset. As discussed in the main paper, noisy segmentation is the most common and detrimental issue, particularly in ScanNet reconstructions, where noisy image and depth data reduce overall scene quality. Filtering these noisy regions results in non-negligible performance gains. Addressing both over-segmentation and under-segmentation also leads to measurable improvements in VQA accuracy, albeit with smaller margins.

We consider all three types of errors to remain a source of imperfection in our pipeline, with our filtering process not being foolproof. As our work does not aim to solve panoptic segmentation, we view the current filtering step, and by extension our full 3D segmentation process, as a component that remains open to further improvement.

\begin{table}[tb!]
\resizebox{\columnwidth}{!}{%
\begin{tabular}{c c c cc}
\hline

Noise Filtering  & Under-Segmentation   & Over-Segmentation & CiDER & EM@1-R   \\ \hline
  \ding{55}       & \ding{55}            &   \ding{55}      &  91.68   &  42.57  \\

 \ding{51}       & \ding{55}            &   \ding{55}      & 94.94     &   43.62   \\
  \ding{51}     & \ding{51}            &     \ding{55}     &   96.25    & 44.04      \\ 
\ding{51}       & \ding{51}      &    \ding{51}     &    97.82 & 46.24             \\ \hline

\end{tabular}%
}
\caption{Performances on VQA (ScanQA subset) with and without filtering.}
\label{tab:vqa-filtering}
\end{table}

\paragraph{Feature Extraction.}
We evaluate the impact of object feature ordering and multi-scale extraction in Table \ref{tab:vqa-extract}. 

Regarding the radar-inspired ordering strategy, we observe that arranging the features such that spatially close objects are also close in the prompt consistently improves performance, regardless of whether multi-scale extraction is used.

The multi-scale experiments further demonstrate that sampling features from each sub-region (i.e., fine- and medium-scale parts) significantly outperforms mono-scale extraction. The “scale fixed size” parameter controls whether a fixed number of features is assigned per object (“Large” scale) or per sub-class/region (“Small” scale). In other words, this setting determines whether all objects contribute equally or proportionally to their internal complexity. Results indicate an advantage in fixing the number of features per object. This may be due to LLaVA's original training, which has been optimized to receive images which contain the same number of features each (i.e. $27\times27$, their number of patches).

\begin{table}[tb!]
\resizebox{\columnwidth}{!}{%
\begin{tabular}{c c c cc}
\hline

Radar Sweeping & Multi-Scale    & Scale Fixed Size & CiDER & EM@1-R   \\ \hline
  \ding{55}       & \ding{55}            &   -       &  82.19    &  38.04    \\

 \ding{51}       & \ding{55}            &   -       &   87.67   & 41.91     \\
   \ding{55}     & \ding{51}            &     Large       &  88.89    & 40.46     \\
   \ding{51}       & \ding{51}      &    Small     &   92.76   &  43.35           \\
  \ding{51}     & \ding{51}            &     Large     &     97.82 & 46.24     \\  \hline

\end{tabular}%
}
\caption{Performances on VQA (ScanQA subset) with or without feature ordering and multi-scale feature extraction.}
\label{tab:vqa-extract}
\end{table}

\begin{figure}[tb!]
      \includegraphics[width=\columnwidth]{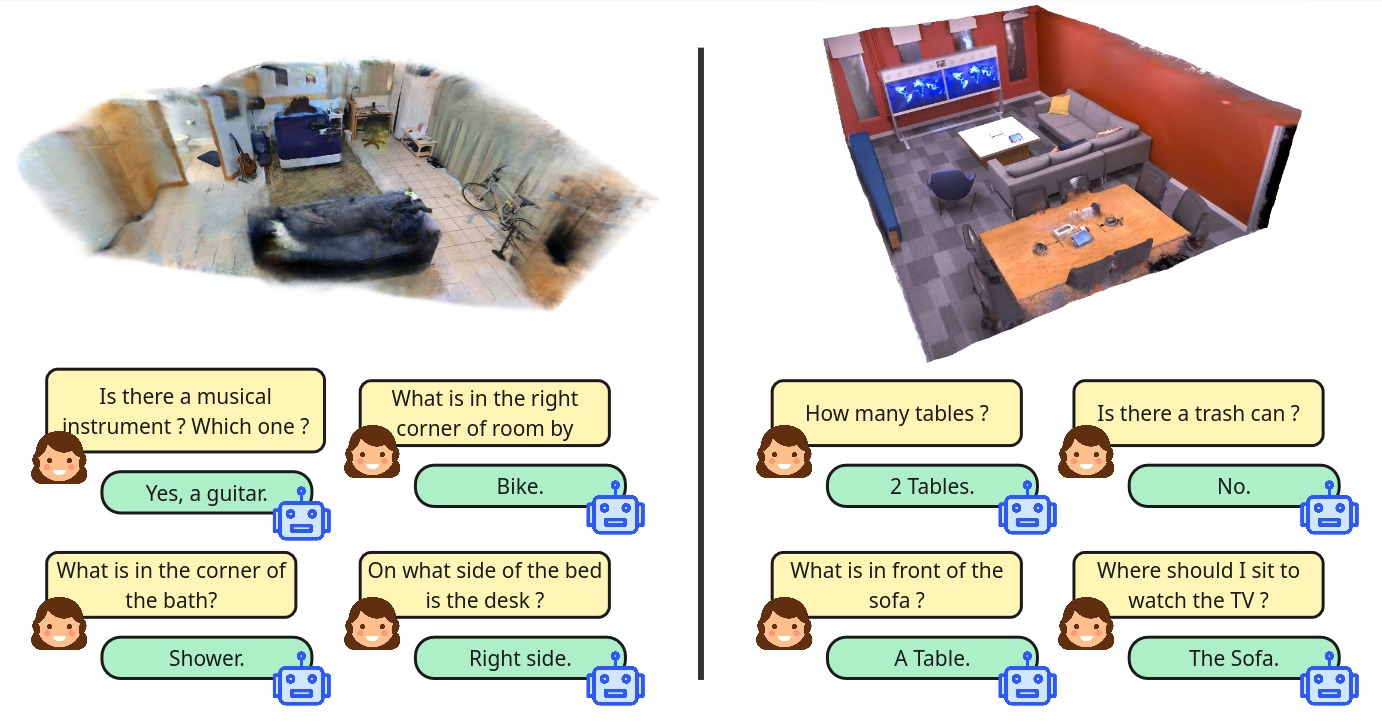}    
    \caption{Further Visual Question Answering Examples.}
    \label{fig:sup-vqa1}
\end{figure}

\paragraph{Number of Features.}
We also evaluate the impact of filling the context window with a different number of visual tokens. SplatTalk analysed that using the equivalent of a single image’s worth of visual tokens produced results comparable to fully filling the context window. Following a similar setup, as reported in Table \ref{tab:vqa-nb-feature}, we observe that while the number of features does have some impact, the effect is indeed relatively minor, compared to our other ablatives. This suggests a high degree of redundancy in the extracted features, and that the VLM can understand the scene even with significantly fewer tokens. Nevertheless, the best performance is still achieved when the context window is fully utilized, likely due to the inclusion of fine-grained cues that aid in more detailed reasoning.

\begin{table}[tb!]
\resizebox{0.8\columnwidth}{!}{%
\begin{tabular}{c cc}
\hline

Number of Total Features & CiDER & EM@1-R   \\ \hline
1000   & 93.19     & 43.77            \\
5000    &  93.34    &  44.35    \\
10000    &  96.26    &  44.35     \\
20000   &   97.01   &    45.51  \\ 
30000   &            97.82 & 46.24\\ \hline

\end{tabular}%
}
\caption{Performances on VQA (ScanQA subset) with varying number of total features.}
\label{tab:vqa-nb-feature}
\end{table}

\begin{figure}[tb!]
\centering
    \includegraphics[width=0.7\columnwidth]{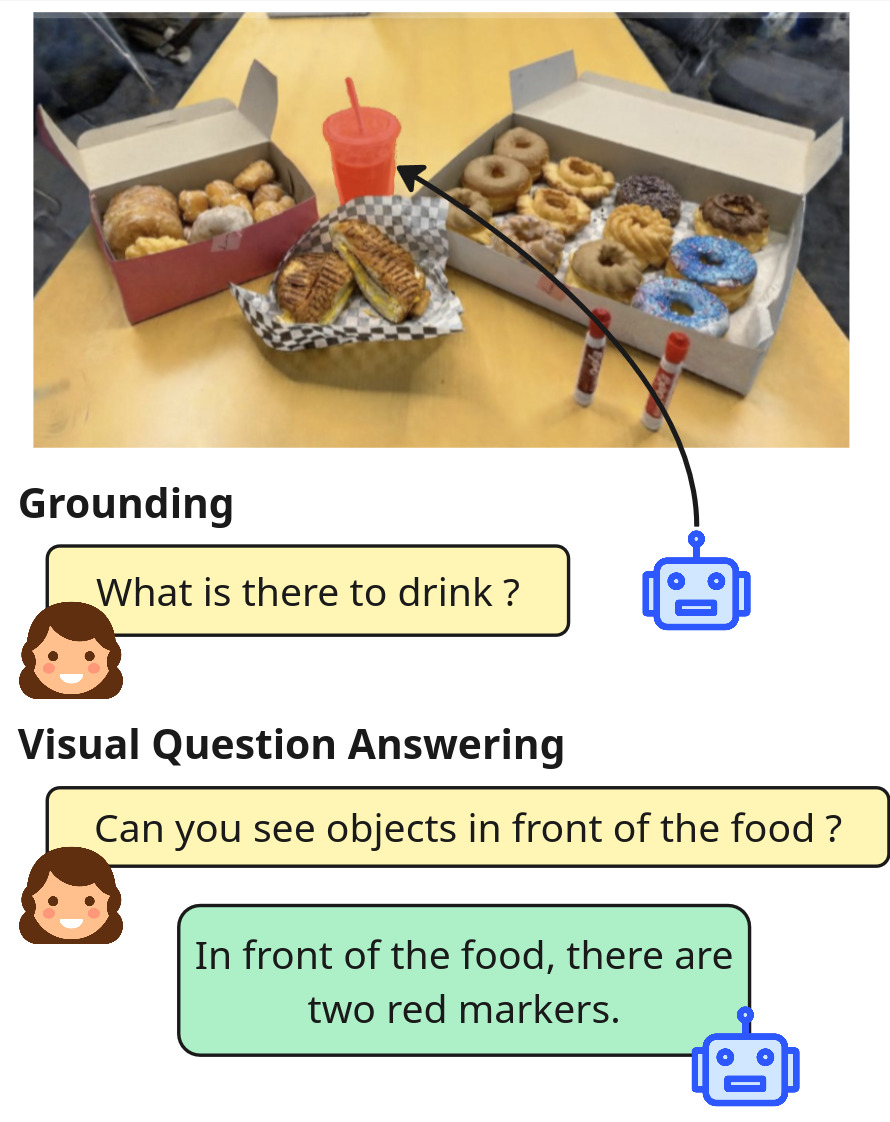}  
    \caption{Grounding and VQA example on the \textit{Donuts} in-the-wild scene.}
    \label{fig:sup-donuts}
\end{figure}

\subsection{Comprehensive 3D Scene Understanding.}
Here, we display additional qualitative results on different downstream tasks associated to scene understanding.

\begin{figure*}[tb!]
    \includegraphics[width=\textwidth]{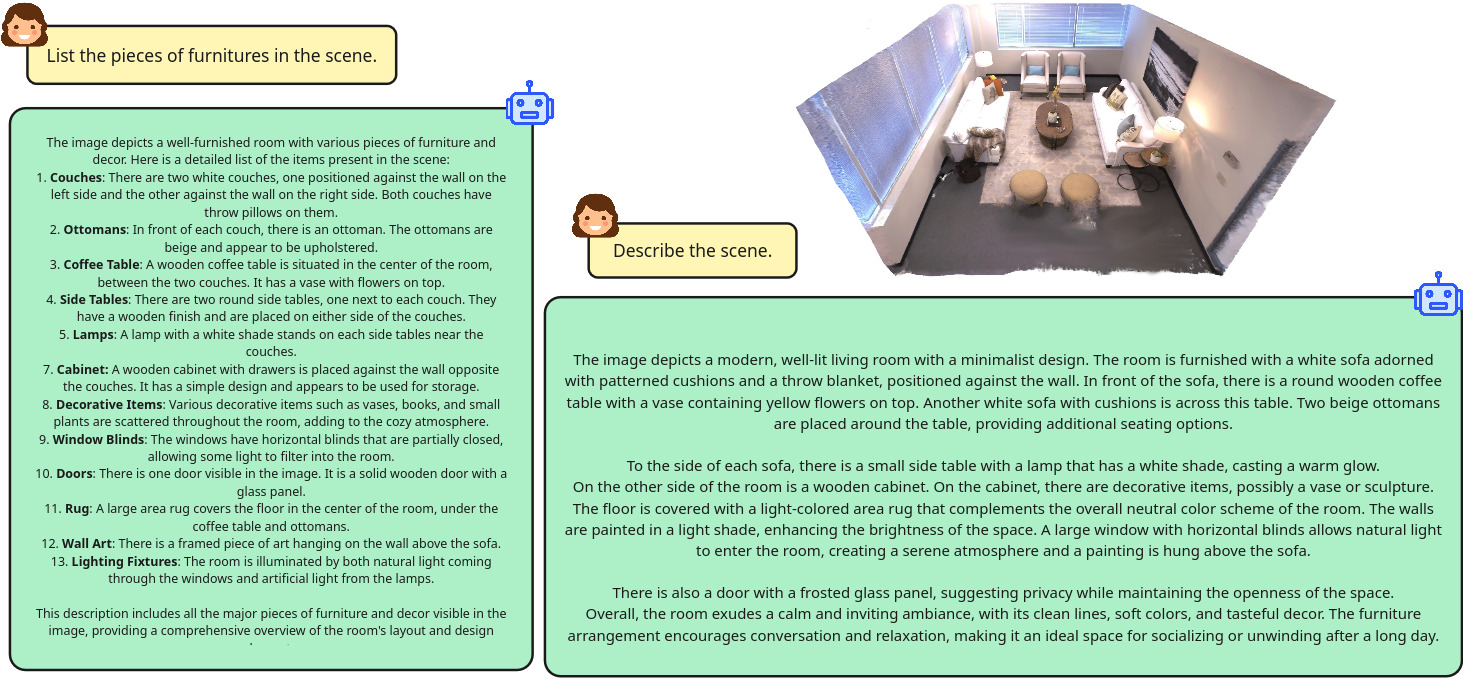}    
    \caption{Open Visual Question Answering Examples.}
    \label{fig:sup-vqa2}
\end{figure*}

\paragraph{VQA and Captioning.}
Figures~\ref{fig:sup-vqa1} and~\ref{fig:sup-vqa2} provide additional examples of visual question answering (VQA) on both Replica and ScanNet scenes, with a focus on open-ended queries. These examples illustrate the capacity of our method to reason about spatial relationships, object existence, counting, and other complex scene-level concepts. Importantly, since our approach does not rely on large-scale training specifically on ScanNet, it demonstrates strong generalization capabilities to novel environments. To highlight this, Figure~\ref{fig:sup-donuts} presents a VQA-Grounding example on an in-the-wild scene from the LeRF dataset ("Donuts"), further showcasing the broad applicability of our framework. 

\begin{figure}[tb!]
    \includegraphics[width=\columnwidth]{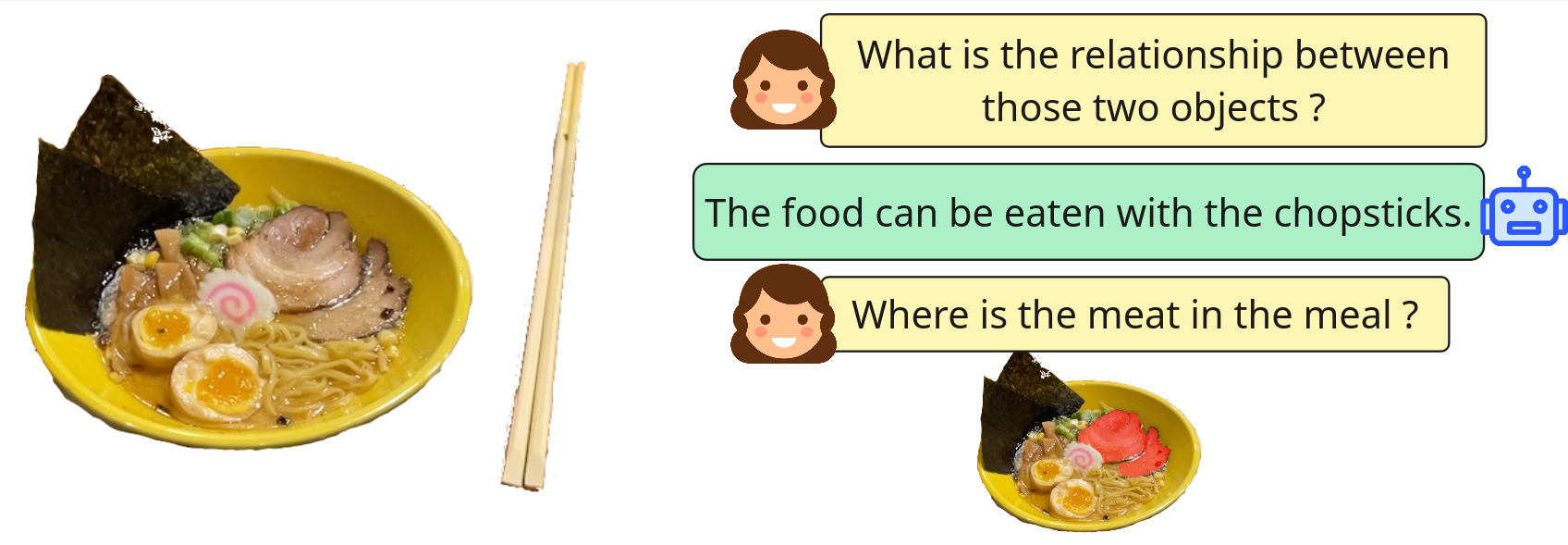}    
    \caption{Example of Object-centric VQA and sub-scale grounding.}
    \label{fig:object-centric}
\end{figure}

We also showcase two examples of object-centric VQA and grounding in Figure~\ref{fig:object-centric} in an in-the-wild scene, further demonstrating the capabilities to perform localized tasks on specific objects.


\begin{figure}[tb!]
    \includegraphics[width=\columnwidth]{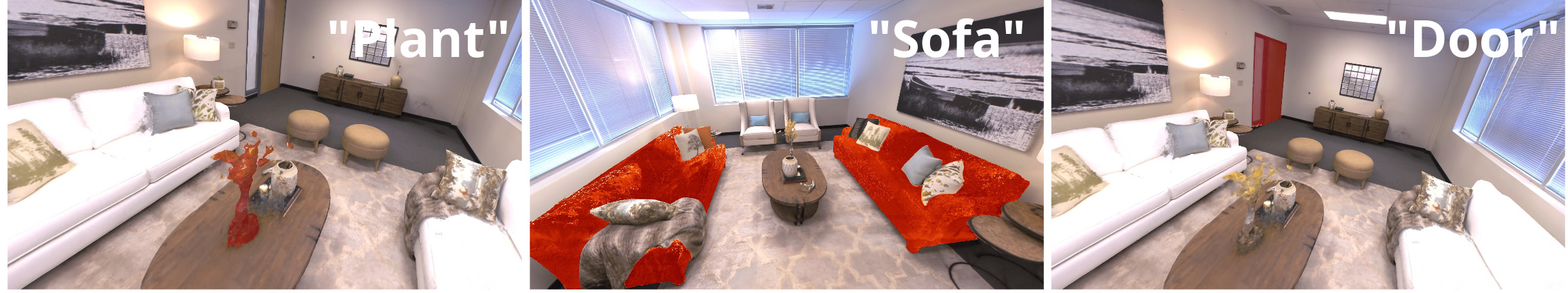}    
    \caption{Some Open-Vocabulary Segmentation examples on Replica's \textit{room0}, using the SAM and CLIP feature field.}
    \label{fig:sup-ovseg}
\end{figure}

\begin{figure}[tb!]
    \includegraphics[width=\columnwidth]{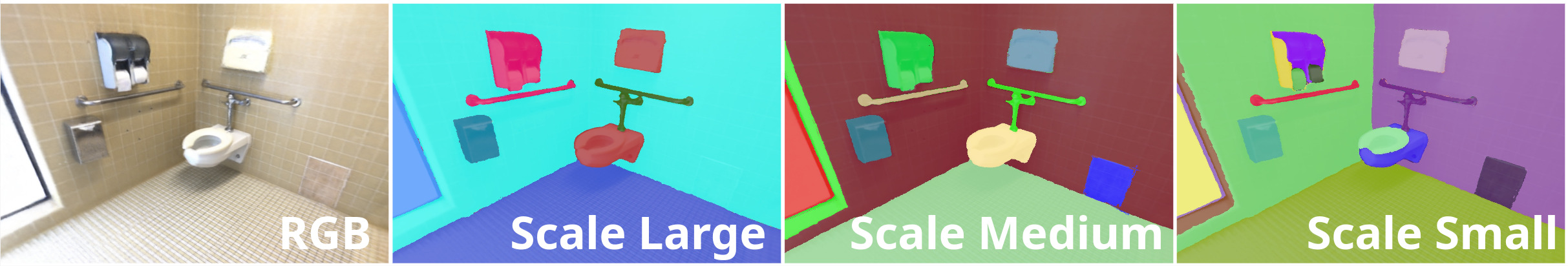}    
    \caption{Multi-Scale Instance Segmentation on ScanNet's \textit{scene0086}.}
    \label{fig:sup-inst}
\end{figure}

\begin{figure}[tb!]
    \includegraphics[width=\columnwidth]{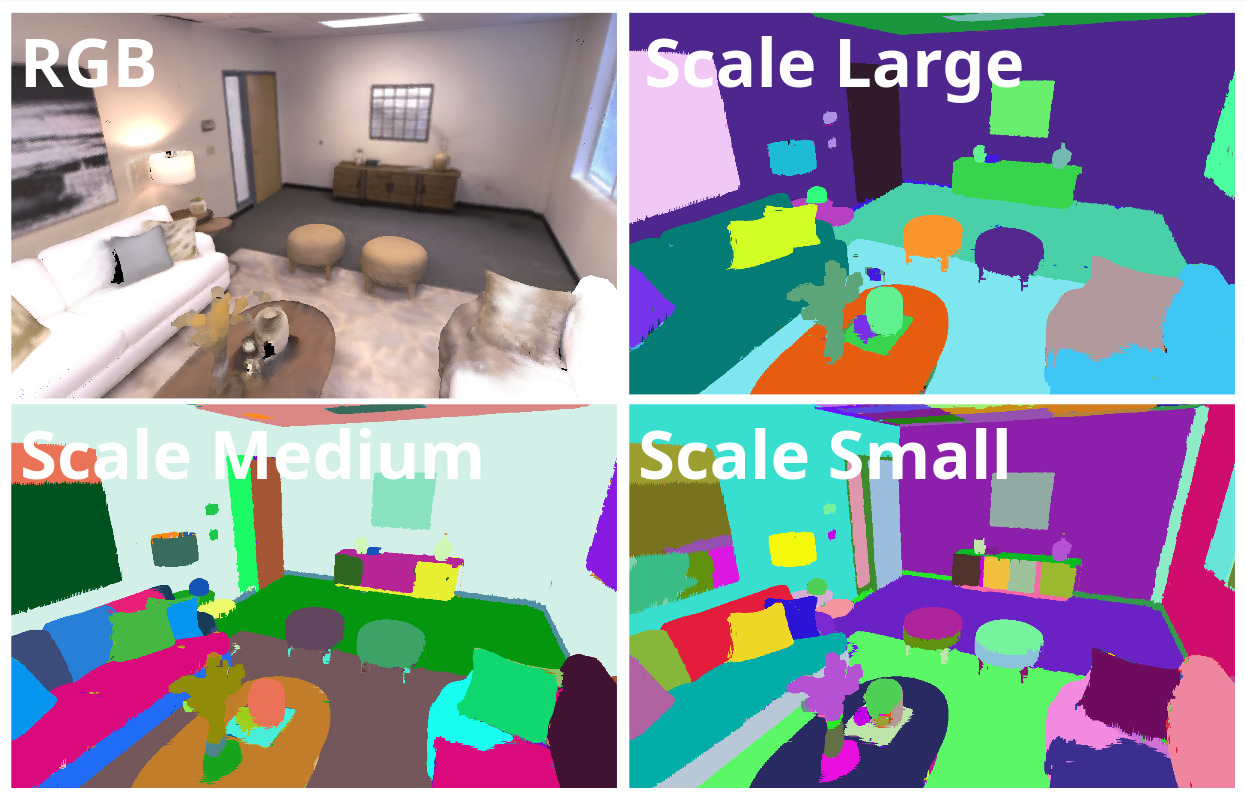}    
    \caption{Multi-Scale Instance Segmentation on Replica's \textit{room0}.}
    \label{fig:sup-instrep}
\end{figure}

\paragraph{Comprehensive Segmentation.}
Figure~\ref{fig:sup-ovseg} presents qualitative examples of open-vocabulary segmentation, a downstream task that is naturally supported by our method. As with semantic segmentation, we follow a similar protocol: a textual query is provided, and segmentation is performed using only the SAM-CLIP feature field, without relying on the LLaVA feature field or any LLM-based reasoning.

In addition, Figure~\ref{fig:sup-inst} showcases a ScanNet example of our multi-scale instance segmentation, demonstrating the ability of our approach to segment a scene into object instances of varying granularities. We also include a complementary example on a Replica scene in Figure~\ref{fig:sup-instrep}, further highlighting the versatility and adaptability of our multi-scale decomposition framework.

\clearpage

\end{document}